\documentclass[journal]{IEEEtran}
\usepackage{xcolor}
\usepackage{mathtools}
\usepackage{bbm}
\usepackage{amsfonts}
\usepackage{multirow}
\usepackage{colortbl}
\usepackage{hyperref}

\ifCLASSINFOpdf
\else
\fi

\usepackage{graphicx}
\usepackage{cite}

\begin{document}
%
\title{Uncertainty-Aware Evaluation for Vision-Language Models}
%
%
%

\author{Vasily Kostumov,
        Bulat Nutfullin,
        Oleg Pilipenko,
        Eugene Ilyushin \\
        \href{https://ensec.ai/}{EnSec-AI} \\
        \{v.kostumov, b.nutfullin, o.pilipenko,
        e.ilyushin\}@ensec.ai 
}

\maketitle

\begin{abstract}
Vision-Language Models like GPT-4, LLaVA, and CogVLM have surged in popularity recently due to their impressive performance in several vision-language tasks. Current evaluation methods, however, overlook an essential component: uncertainty, which is crucial for a comprehensive assessment of VLMs. Addressing this oversight, we present a benchmark incorporating uncertainty quantification into evaluating VLMs. 

Our analysis spans 20+ VLMs, focusing on the multiple-choice Visual Question Answering (VQA) task. We examine models on 5 datasets that evaluate various vision-language capabilities.

Using conformal prediction as an uncertainty estimation approach, we demonstrate that the models' uncertainty is not aligned with their accuracy. Specifically, we show that models with the highest accuracy may also have the highest uncertainty, which confirms the importance of measuring it for VLMs.
Our empirical findings also reveal a correlation between model uncertainty and its language model part.  

The code is available at \url{https://github.com/EnSec-AI/VLM-Uncertainty-Bench}.

\end{abstract}


%
\IEEEpeerreviewmaketitle

\section{Introduction}

In the fast-paced field of artificial intelligence, significant advancements have been achieved by large language models (LLMs) like GPT-4 \cite{GPT4}, LLaMA \cite{LLAMA}, and LLaMA2 \cite{LLAMA2} in both natural language understanding (NLU) and generation (NLG). To exploit these NLU and NLG capabilities for tasks that combine vision and language, a common strategy involves augmenting LLMs with visual data as an additional input form and synchronizing this data with textual information. This technique has been successfully implemented in various Vision-Language Models (VLMs), including MiniGPT-4 \cite{MINIGPT4}, LLaVA \cite{LLAVA}, LLaVA-1.5 \cite{LLAVA15}, among many other Vision-Language Models.

\begin{figure}
  \centering
  \includegraphics[width=3.5in]{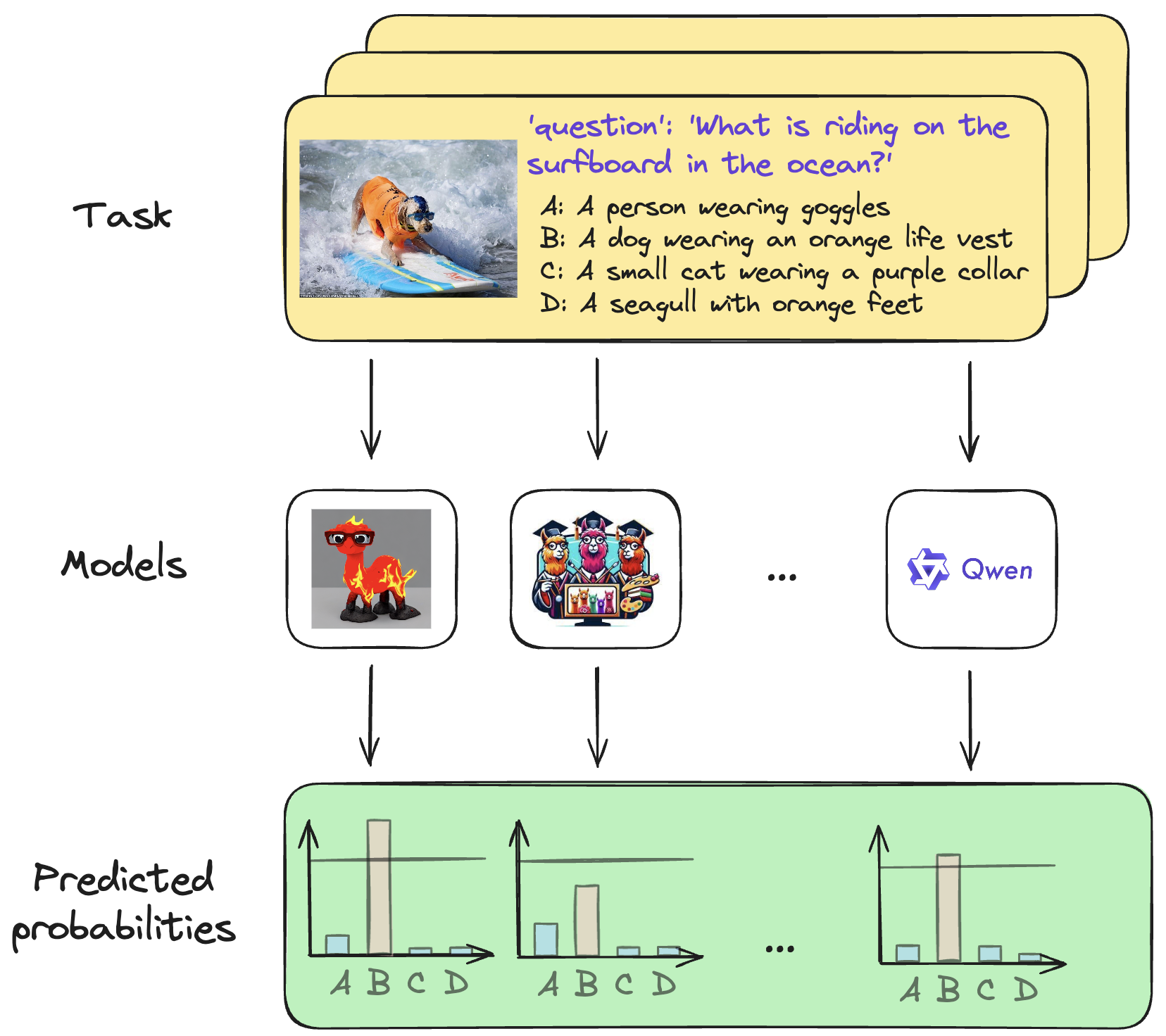}
  \caption{Several VLMs predict correct answers while demonstrating different levels of certainty. In case of an incorrect answer, the models' confidence may also vary.}
  \label{fig:logo}
\end{figure}

However, conducting a comprehensive evaluation of VLMs remains
a challenging endeavor. VLM benchmarks, including VQAv2 \cite{VQAv2}, GQA \cite{GQA}, MMBench\cite{mmbench}, MMMU\cite{mmmu}, ScienceQA\cite{scienceqa}, and SEEDBench\cite{seedbench}, provide a foundation for evaluating the multifaceted capabilities of VLMs. These benchmarks despite their usefulness, possess a significant
limitation: they do not take into account the uncertainty of models.

Comprehensively assessing VLMs extends beyond performance metrics, delving into critical aspects such as security, ethics, fairness, and robustness. Ensuring VLMs adhere to ethical guidelines, exhibit fairness across diverse demographics, and maintain robustness against adversarial attacks are paramount for their safe and equitable application. Furthermore, the evaluation of VLMs must also consider their susceptibility to biases, potential for misuse, and implications on privacy. This comprehensive approach underscores the complexity of VLM evaluation, highlighting the need for a holistic understanding of their impacts across various dimensions.

Nevertheless, as illustrated in Figure \ref{fig:logo}, while two VLMs can achieve the same accuracies, they might demonstrate different levels of uncertainty about each question. This situation can be likened to students answering multiple-choice questions on an exam, where two students might choose the identical answer but have differing levels of confidence or understanding regarding the question. Hence, integrating uncertainty into the evaluation framework is crucial for a fuller appraisal of LLMs.

In this paper, we introduce the application of conformal prediction \cite{Vovk, angelopoulos2022gentle} as a technique to measure uncertainty in VLMs, positioning it as a practical and theoretically sound approach for evaluating VLMs' uncertainty. We follow the approach from \cite{ye2024benchmarking}, who applied a conformal prediction framework for benchmarking Large Language Models uncertainty jointly with their accuracy.
Additionally, we have incorporated supplementary metrics, including the Expected Calibration Error (ECE), and conducted evaluations on 20 VLMs, facilitating a comprehensive comparison. 

Our main contribution can be summarized as follows:
\begin{itemize}
    \item We prepare five various datasets for Visual Question Answering and unify them for our benchmark
    \item We extensively examine 9 Vision-Language Models series in terms of their uncertainty and accuracy. Specifically, we use a conformal prediction approach and Expected Calibration Error to estimate uncertainty
    \item Our experiments reveal that accuracy and uncertainty are not aligned as far as models with higher accuracy may have higher uncertainty. Also, we show that with increasing LLM size, the uncertainty of VLM decreases
\end{itemize}

\section{Related works}

\subsection{Multimodal Large Language Models}

\begin{figure*}[h!]
  \centering
  \includegraphics[width=0.8\linewidth]{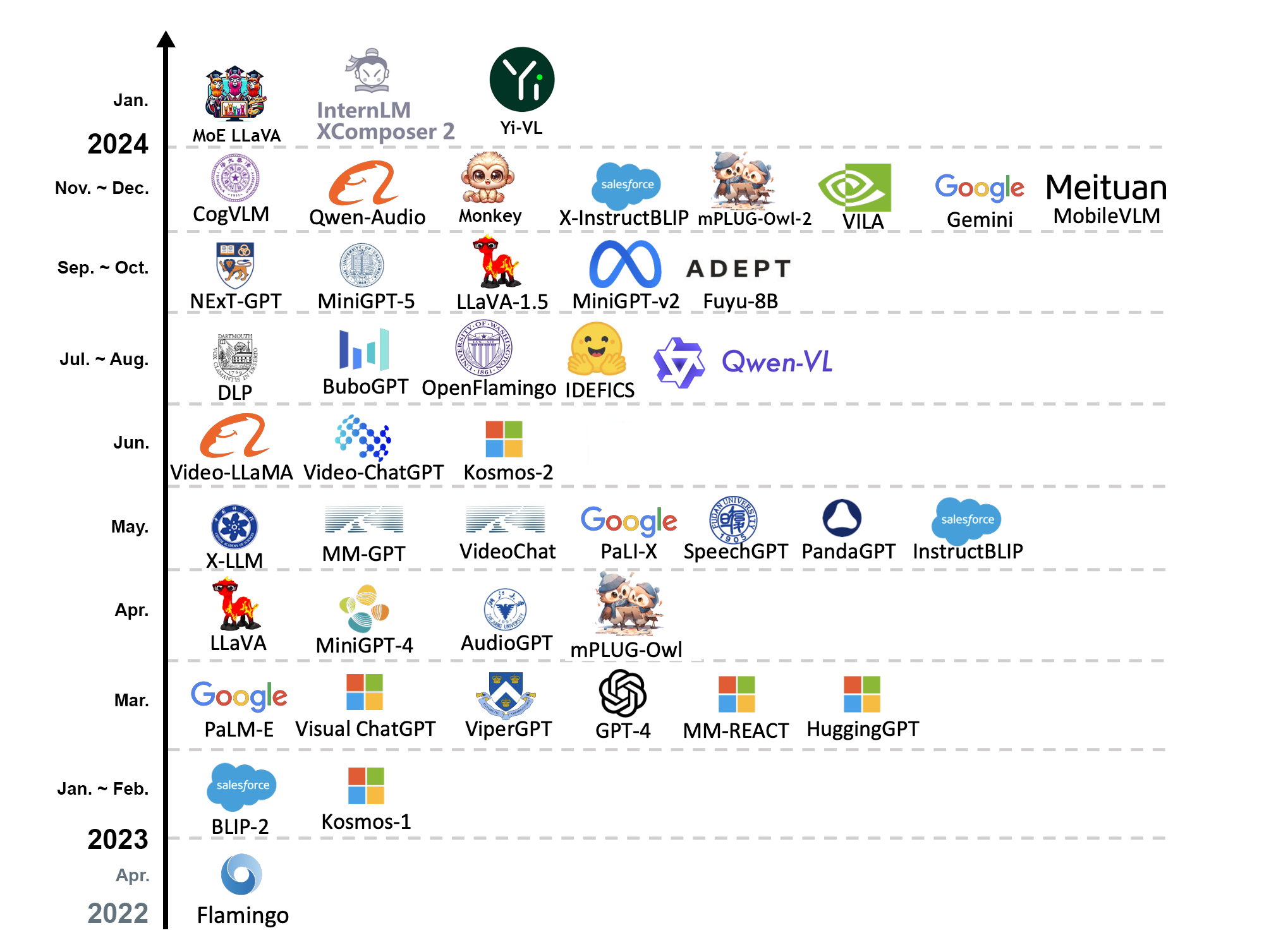}
  \caption{Timeline of MultiModal Large Language Models (inspired by \cite{zhang2024mmllms})}
  \label{fig:timeline}
\end{figure*}
The timeline of Multi-Modal Large Language Models is presented in Figure \ref{fig:timeline}.

The emergence of GPT-4 \cite{GPT4} and Gemini \cite{geminiteam2023gemini} has led to a remarkable breakthrough in the development of large language models (LLMs). These cutting-edge models have demonstrated unparalleled abilities in multi-modal comprehension and generation, paving the way for future more advanced and sophisticated language processing systems. This breakthrough ignited a wave of research enthusiasm toward the development of advanced LLMs, initiating a series of innovations starting from Flamingo \cite{awadalla2023openflamingo}, BLIP2 \cite{li2023blip2}, and Kosmos-1 \cite{huang2023language}, extending through Fuyu-8B \cite{fuyu-8b}, OtterHD-8B \cite{li2023otterhd} and exploration of audio understanding with models like QwenAudio \cite{chu2023qwen}.

Simultaneously, notable progress has been achieved in creating specific modalities. Kosmos-2 \cite{peng2023kosmos2} and MiniGPT-5 \cite{zheng2023minigpt5} have advanced image generation, and SpeechGPT \cite{zhang2023speechgpt} has improved speech generation.

Recent research has been geared towards achieving human-like capability for any-to-any modality conversion. Projects like Visual-ChatGPT \cite{wu2023visual}, ViperGPT \cite{suris2023vipergpt}, MMREACT \cite{yang2023mmreact}, HuggingGPT \cite{shen2023hugginggpt}, and AudioGPT \cite{huang2023audiogpt} are integrating LLMs with tools for enhanced multi-modal generation.

Addressing the challenge of error propagation in cascaded systems, novel approaches like NExT-GPT \cite{wu2023nextgpt} and CoDi-2 \cite{tang2023codi2} have been introduced, offering end-to-end multi-modal LLM solutions capable of handling arbitrary modalities. These advancements underscore the rapid progress and diverse exploration within the field, highlighting the continuous effort to bridge the gap between human and machine understanding across multiple modalities.

Visual Language Models have gained significant popularity, beginning with Kosmos-1 \cite{huang2023language} and extending to Fuyu-8B \cite{fuyu-8b}. Their capabilities have been notably enhanced through methodologies such as those introduced by LLaVA \cite{LLAVA} and LLaVA-1.5 \cite{LLAVA15}. Currently, the existence of numerous models and their respective fine-tunings indicates rapid growth in this particular subfield. 

\subsection{Uncertainty estimation}

Comprehending and measuring uncertainty in machine learning algorithms is essential due to its importance in autonomous driving and real-world applications such as healthcare, decision-making, and risk assessment.  

Methods for uncertainty quantification can be divided into four categories \cite{gawlikowski2021survey}:
single deterministic methods, ensemble methods,
Bayesian methods, and test-time augmentation methods.
Single deterministic methods quantify prediction uncertainty based
on only one forward pass and designed for deterministic model. Ensemble
methods estimate uncertainty based on a set of different ML model outputs. Bayesian methods only leverage the inherent
randomness of a model. Test-time augmentation methods augment the input data during evaluation to measure a model's prediction uncertainty. The last three categories are computationally expensive in practice and can hardly be used for large models.

A significant part of the single deterministic methods is based on predicted confidence. However, to make use of uncertainty quantification methods, one has to be sure that the model is well-calibrated \cite{gawlikowski2021survey}, which means that derived predictive confidence represents a good approximation of the actual
probability of correctness. 

To estimate model's calibration, different calibration measures can be considered. A widely utilized technique, Expected Calibration Error (ECE), evaluates the discrepancy between a model's predicted probabilities and the actual frequencies of correct predictions \cite{calibration}. Despite its widespread use, this metric has no formal guarantees in terms of calibration errors. The same limitations are inherent in many other similar methods.

Recently, there has been an increasing interest in the application of conformal prediction \cite{Vovk} \cite{Fontana} for robust uncertainty quantification. This framework has also found wide usage in natural language processing tasks, including paraphrase detection \cite{giovannotti2021transformer}, part-of-speech prediction \cite{dey2022conformal}, sentiment analysis \cite{maltoudoglou2020bert} and language modeling \cite{quach2023conformal}. 

 Conformal prediction facilitates the identification of a subset of plausible labels in classification tasks, offering a statistically valid measure of confidence regarding the true label's inclusion. The magnitude of this subset effectively quantifies the degree of uncertainty, rendering conformal prediction an advantageous method for uncertainty quantification. 
 The advantage of this approach is that it provides a statistically robust coverage guarantee.

Recently, several works \cite{kumar2023conformal, ye2024benchmarking} utilized conformal prediction for uncertainty estimation in multiple-choice question answering for LLMs. While in \cite{kumar2023conformal} MMLU benchmark \cite{hendrycks2020measurng} was employed to quantify the LLM uncertainty, in \cite{ye2024benchmarking} uncertainty was estimated across five different tasks.

\section{Vision-Language Model uncertainty estimation}

In this section we give some base about formal statement for VLMs and Uncertainty Estimation.

\subsection{Vision-Language Models task}

Given an image \( I \) and a text fragment \( T \), the task is to predict the next token \( y \). The visual language model utilizes a projector \( P \) to map the encoded representation of the image into the word embedding sequence. This process can be formally described as follows:

\begin{equation}
\theta(x^T, x^I) = \theta_{dec}\left(\theta_{emb}^T(x^T) \parallel P(\phi_{enc}^I(x^I))\right)
\end{equation}

Here, \( x^T \) denotes the encoded representation of text, \( x^I \) represents the encoded image, \( \theta_{emb}^T \) is the text embedding function, \( \phi_{enc}^I \) is the image encoder function, \( P \) is the projector that maps the image encoding to the word embedding space, \( \parallel \) signifies concatenation, and \( \theta_{dec} \) is the decoder function that generates the prediction for the next token \( y \).

\subsection{Adapters}
\subsubsection{Visual Encoder}
In the visual modality for image processing, four distinct architectures are highlighted: ViT, CLIP ViT, and Eva-CLIP ViT. ViT transforms images into a series of patches for processing with Transformer blocks, applying the Transformer architecture directly to visual data. CLIP ViT integrates text and image understanding by using a ViT and a text encoder optimized through contrastive learning from a large corpus of text-image pairs. Eva-CLIP ViT further refines this approach by stabilizing the training and optimization of CLIP.

In Fuyu-8B\cite{fuyu-8b} architecture
image patches are linearly projected directly into the first layer of the transformer, bypassing the embedding lookup. This simplified architecture supports arbitrary image resolutions and simplifies both training and inference.

\subsubsection{Input Projector}
The projector's $P$ task is to map encoded features $\phi_{enc}^I(x^I)$ into the textual feature space.

Multimodal encoded features projected into textual feature space $P(\phi_{enc}^I(x^I)))$, combined with the textual features $\theta_{emb}^T(x^T)$, are fed into the LLM part.

To implement the Input Projector, straightforward approaches like a Linear Projector or a Multi-Layer Perceptron (MLP) which consists of multiple linear layers separated by nonlinear activations, are used. 

More sophisticated methods include Cross-attention, Q-Former \cite{li2023blip2}, or P-Former \cite{jian2023bootstrapping}. The Cross-attention mechanism compresses the sequence of features $F_X$ into a fixed length using trainable query vectors and these features as keys, enabling direct input into the LLM \cite{bai2023qwenvl}.

\subsection{Models}\label{par:models}

Here, we briefly describe the models we use for comparison in our benchmark.

The LLaVA model \cite{LLAVA} employs a pre-trained CLIP ViT to encode images, which are then passed through a linear projector into the word embedding space.

LLaVA-1.5 \cite{LLAVA15} replaces the linear projector with a two-layer MLP and improves upon its predecessor by increasing the image resolution to 336 pixels. Both versions utilize the LLAMA LLM architecture \cite{LLAMA} \cite{LLAMA2}.

CogVLM\cite{Cogvlm} utilizes a ViT encoder, an MLP projector, and the Vicuna1.5-7B LLM. A visual expert module, comprising a QKV matrix and MLP, is added to each layer to facilitate deep visual-language feature alignment. These modules are shaped and initialized in accordance with the pre-trained language model.

Yi-VL adopts the LLaVA1.5 architecture, and its LLM is initialized with Yi-34B-Chat or Yi-6B-Chat.

Qwen-VL combines the pre-trained Qwen-7B LLM \cite{qwen} with an OpenCLIP ViT image encoder. It utilizes a single-layer cross-attention mechanism with trainable query vectors and image feature keys to compress the feature sequence to a length of 256. This adapter also integrates 2D absolute positional encodings within the cross-attention to maintain positional integrity in the compressed features before they are fed into the LLM.

Monkey \cite{li2023monkey} features a higher resolution of up to 1344×896 pixels without the need for pre-training, surpassing the conventional 448×448 resolution. It processes images in segments to extract local features and operates a separate branch for global features. This model uses the ViT-BigHuge and the LLM from QwenVL \cite{qwen}.

mPLUG-Owl \cite{ye2023mplugowl} uses a ViT-L and the LLaMA-7B \cite{LLAMA} as LLM. It introduces a visual abstractor module with cross-attention that condenses visual information into several learnable tokens, resulting in higher semantic visual representations with reduced computational requirements.

InternLM-XComposer-VL  \cite{zhang2023internlmxcomposer} uses EVA-CLIP, an advanced version of the standard CLIP model with additional masked image modeling features. Its perceive sampler, using attentive pooling, reduces 257 image tokens to 64, aligning them with the large language model by leveraging a BERTbase model with cross-attention layers. This model incorporates InternLM-Chat-7B \cite{team2023internlm}.

MoE-LLaVA \cite{lin2024moellava} employs a CLIP-Large as the image encoder. Its MLP projector consists of two linear layers with a GELU activation function and utilizes a novel mixture of experts in the LLaMA architecture.

\subsection{Conformal Prediction} 

\textbf{Prediction Sets}.
The main idea behind conformal prediction \cite{Vovk, angelopoulos2021gentle, Fontana} is to produce prediction regions or sets rather than point estimates. This prediction set consisting of possible labels and including the correct label with a predefined error rate naturally encodes the model's uncertainty about any particular input by
the size of the prediction set.

Formally, let $f: \mathcal{X} \rightarrow \mathcal{Y}$ be a classifier that outputs softmax score  $ \Delta^{\lvert \mathcal{Y} \rvert}$ and class with maximum score is then predicted. For any unseen test input $X_{test}$ conformal prediction procedure predict set of labels $\mathcal{C}(X_{test}) \in  2^\mathcal{Y}$ such that :
\begin{equation}
    1 - \alpha \leq \mathbb{P}\left(Y_{test} \in \, \mathcal{C}(X_{test})\right),
\end{equation}
where $\alpha \in (0, 1)$ is the predefined error rate  and $Y_{test}$ is the correct label of $X_{test}$. \\
\textbf{Score functions.}
To calculate prediction sets score function $S(X, Y) \in \mathbb{R}$ is utilized. 
The softmax score is not an appropriate choice for a conformal score function since modern neural networks are known to be over-confident. Two conformal score functions using softmax scores are commonly employed instead.

\begin{itemize}
    \item \textbf{Least Ambiguous set-valued Classifiers (LAC)} \cite{sadinle2019least}.  
    \begin{equation}
        \label{eq:lac}
        S(X, Y) = 1 - f_{Y}(X), 
    \end{equation}
    where $f_{Y}(X)$ is the softmax score corresponding to the ground truth label $Y$
    \item \textbf{Adaptive Prediction Sets (APS)} 
    \cite{romano2020classification}
    \begin{equation} 
        \label{eq:aps}
        S(X, Y) = \sum_{\{j \in \mathcal{Y}: f_j(X) \ge f_Y(X)\}} f_j(X),
    \end{equation}
    where $f_j(X)$ is the softmax score corresponding to label $j$. Thus, this score sums softmax scores, which are greater or equal to softmax scores for ground truth label.
    
\end{itemize} 
\textbf{Calibration.}
To fulfill the coverage guarantee requirement for a predefined error rate $\alpha$ on a test set, we need to calibrate the prediction sets for the selected score function. Special calibration set  $\mathcal{D}_{cal} = \{(X^{(i)}_{cal}, Y^{(i)}_{cal})\}_{i=1}^{n}$ are used for that purpose. 

For every pair from the calibration set, the score function needs to be calculated $s_i = S(X^{(i)}_{cal}, Y^{(i)}_{cal})$. After that we estimate a threshold $\hat{q}$ as a $ \lceil(n+1)(1-\alpha)\rceil / n$ quantile of this set:
\begin{equation}
    \hat{q} = \text{quantile}\left(\{s_1, \ldots, s_n\}, \frac{\lceil{(n+1)(1-\alpha)\rceil}}{n}\right),
\end{equation} \\
\textbf{Inference.}
For each input $X$ in  $\mathcal{D}_{test}$ and for every $y \in \mathcal{Y}$ prediction sets can be constructed as follows:
\begin{equation}
    \mathcal{C}(X) = \left\{y \in \mathcal{Y}: S(X_{test}, y) \leq \hat{q} \right\}
\end{equation} \\
\textbf{Benefits.} The main benefits of conformal prediction, besides the guarantees mentioned above, are:
\begin{itemize}
    \item Conformal prediction is model-agnostic. It doesn't rely on specific assumptions about the model's inner workings. This could be used for the pre-trained model, and only logits/softmax scores are needed. 
    \item Conformal prediction is distribution-free. It doesn't require specific assumptions about the underlying data distribution. 
    \item Interpretability. Prediction sets are interpretable in the sense that they reflect the set of options that the model considers.
\end{itemize}

\section{Evaluation Tasks and Datasets}

Evaluating the performance of vision language models on multiple benchmarks related to different tasks is standard practice. These benchmarks can be categorized into three main tasks \cite{Cogvlm}:  Image Captioning, Visual Question Answering, and Visual Grounding. 

The first (NoCaps \cite{Nocaps}, COCO\cite{Coco}, Flickr30K \cite{Flickr30K}) evaluates the ability of the models to generate textual captions for a given image. The second contains 
benchmarks with images and questions to them. 
The task of the model is to predict correct answer without options (\textbf{open-ended VQA}: VQAv2 \cite{VQAv2}, GQA \cite{GQA}, TextVQA \cite{TextVQA}, MM-Vet \cite{MM-Vet}) or to choose
correct option from some predefined set (\textbf{multiple-choice VQA}: ScienceQA \cite{scienceqa}, SEED-Bench \cite{seedbench}, MMBench \cite{mmbench}, MMMU
\cite{mmmu}, OODCV-VQA \cite{oodcv}, AI2D \cite{ai2d}, POPE \cite{pope}, MathVista \cite{vista}, MME \cite{mme}). The latter task is to locate the most relevant object or region in an image, based on a natural language query and include RefCOCO/RefCOCO+, RefCOCOg \cite{refcoco} benchmarks.

In this work, we follow \cite{ye2024benchmarking} and prepare 
datasets for multiple-choice question answering (MCQA) task. For that
purpose, we adapt some datasets from the multiple-choice VQA 
above. Some of the datasets from that list are binary classification
task with yes/no answers, and we do not take them into consideration. \\

\textbf{MMBench}\cite{mmbench} comprises around 3000 multiple-choice questions in the test set and 4000 questions in the dev set spanning 20 distinct capability dimensions. MMBench employs a hierarchical structure for capability dimensions, featuring two overarching categories: \textbf{perception and reasoning}. Since the ground truth answer is unavailable for the test set, we use only dev set
in our evaluation. In this set, the number of answer options for each question varies from two to four and we add some randomly sampled incorrect options from other questions to make it four
in every question.  \\

\textbf{OODCV-VQA} \cite{oodcv} is a part of the safety evaluation benchmark also containing Sketchy-VQA dataset and Redteaming Datasets, including Misleading Attack through the ViT and LLM Jailbreaks. We adopt the "Digits" subset of OODCV-VQA as follows. Each question in this subset applies to \textbf{Out-of-Distribution Instance Counting}
task, and the number of options to choose from is two. Since the 
correct answer is always in range from zero to five, we randomly
sample two incorrect options from four digits, which were not included in the options by the authors. \\

\textbf{ScienceQA} \cite{scienceqa} consists of scientific questions and includes three subjects: natural science, language science, and social science. We adopt validation and test part of this benchmark selecting only questions with images and closed choice category. The number of options varies from two to five in the selected questions, and we add some random incorrect answers, like in MMBench when the number of options is less than four or randomly delete one incorrect option when the number of options is five. Finally, we have 3,952 \textbf{Scientific Reasoning} questions with four answer options. \\

\textbf{SEEDBench} \cite{seedbench} spans 12 evaluation dimensions, including spatial and temporal understanding. 
For our benchmark, we selected only dimensions 1-9, which are related to
image modality. This dimension evaluates VLM ability in tasks such as  \textbf{Scene Understanding, Instance Identity, Instance Location} and others. Finally, we utilized 14,233 questions from this benchmark with four answer options. \\

\textbf{AI2D} \cite{ai2d} contains 5,000+ diagrams that cover various topics from elementary school science, encompassing diverse complexities and 15,000+ multiple-choice questions to these diagrams, which test the model ability of \textbf{Diagram Understanding} and reason about the information presented in the diagrams. Each question already has four answer options, so we leave it unchanged. 

Following \cite{ye2024benchmarking}, two additional choices ("I don't know" and "None of the above") were also appended to the list of options, expanding the total options for each question to six.

\begin{table*}[h]
\centering
\resizebox{\textwidth}{!}{%
\begin{tabular}{l|l|cccccc|cccccc}
\hline
\multirow{2}{*}{\textbf{VLM}} & \multirow{2}{*}{\textbf{LLM}} & \multicolumn{6}{c|}{\textbf{Coverage Rate (\%)}} & 
\multicolumn{6}{c}{\textbf{Acc (\%)} $\uparrow$} \\ 
\cline{3-14} 
&  & \textbf{MMB} & \textbf{OOD} & \textbf{SQA} & \textbf{SB} & \multicolumn{1}{c|}{\textbf{AI2D}} & \textbf{Avg.} & \textbf{MMB}  & \textbf{OOD} & \textbf{SQA} & \textbf{SB} & \multicolumn{1}{c|}{\textbf{AI2D}} & \textbf{Avg.} \\ \hline 
LLaVA-v1.6-13B & Vicuna-13b & 94.59 & 93.60 & 94.28 & 93.60 & 94.67 & 94.15 & \cellcolor{red!45} 76.75{\tiny (2)} & \cellcolor{blue!30} 72.93{\tiny (5)} & \cellcolor{red!35} 70.56{\tiny (4)} & \cellcolor{blue!50} 70.37{\tiny (1)} & \multicolumn{1}{>{\columncolor{red!50}}c|} {73.67{\tiny (1)}} & \cellcolor{blue!50} 72.85{\tiny (1)} \\
Monkey-Chat & Qwen-7b & 93.65 & 92.51 & 94.64 & 92.86 & 94.13 & 93.56 & \cellcolor{red!50} 76.98{\tiny (1)} & \cellcolor{blue!25} 70.6{\tiny (6)} & \cellcolor{red!45} 74.66{\tiny (2)} & \cellcolor{blue!20} 66.1{\tiny (7)} & \multicolumn{1}{>{\columncolor{red!40}}c|} {67.95{\tiny (3)}} & \cellcolor{blue!45} 71.26{\tiny (2)} \\
LLaVA-v1.6-7B & Vicuna-7B & 93.86 & 93.50 & 93.85 & 93.47 & 92.92 & 93.52 & \cellcolor{red!35} 75.56{\tiny (4)} & \cellcolor{blue!40} 73.7{\tiny (3)} & \cellcolor{red!20} 65.86{\tiny (7)} & \cellcolor{blue!45} 69.06{\tiny (2)} & \multicolumn{1}{>{\columncolor{red!45}}c|} {69.75{\tiny (2)}} & \cellcolor{blue!40} 70.78{\tiny (3)} \\
InternLM-XComposer2-VL & InternLM-7b & 92.87 & 90.72 & 94.16 & 92.18 & 93.08 & 92.60 & \cellcolor{red!10} 71.77{\tiny (9)} & \cellcolor{blue!20} 70.04{\tiny (7)} & \cellcolor{red!50} 77.95{\tiny (1)} & \cellcolor{blue!15} 64.44{\tiny (8)} & \multicolumn{1}{>{\columncolor{red!35}}c|} {66.13{\tiny (4)}} & \cellcolor{blue!35} 70.07{\tiny (4)} \\
Yi-VL-6B & Yi-6B & 94.43 & 92.69 & 93.96 & 92.89 & 93.89 & 93.57 & \cellcolor{red!30} 75.24{\tiny (5)} & \cellcolor{blue!45} 73.91{\tiny (2)} & \cellcolor{red!25} 66.72{\tiny (6)} & \cellcolor{blue!25} 66.25{\tiny (6)} & \multicolumn{1}{>{\columncolor{red!25}}c|} {58.84{\tiny (6)}} & \cellcolor{blue!30} 68.19{\tiny (5)} \\
CogAgent-VQA & Vicuna-7B & 94.61 & 93.00 & 93.80 & 92.65 & 92.24 & 93.26 & \cellcolor{red!25} 74.78{\tiny (6)} & \cellcolor{blue!15} 68.57{\tiny (8)} & \cellcolor{red!30} 67.12{\tiny (5)} & \cellcolor{blue!40} 68.01{\tiny (3)} & \multicolumn{1}{>{\columncolor{red!20}}c|} {58.2{\tiny (7)}} & \cellcolor{blue!25} 67.34{\tiny (6)} \\
MobileVLMV2-7B & Vicuna-7B & 93.81 & 93.57 & 94.06 & 93.58 & 93.05 & 93.61 & \cellcolor{red!40} 75.97{\tiny (3)} & \cellcolor{blue!10} 66.53{\tiny (9)} & \cellcolor{red!40} 72.33{\tiny (3)} & \cellcolor{blue!35} 66.71{\tiny (4)} & \multicolumn{1}{>{\columncolor{red!5}}c|} {53.55{\tiny (10)}} & \cellcolor{blue!20} 67.02{\tiny (7)} \\
MoE-LLaVA-Phi2-2.7B & Phi2-2.7B & 94.38 & 91.56 & 93.96 & 93.63 & 93.29 & 93.36 & \cellcolor{red!20} 73.73{\tiny (7)} & \cellcolor{blue!50} 74.82{\tiny (1)} & \cellcolor{red!10} 64.04{\tiny (9)} & \cellcolor{blue!30} 66.42{\tiny (5)} & \multicolumn{1}{>{\columncolor{red!15}}c|} {55.76{\tiny (8)}} & \cellcolor{blue!15} 66.95{\tiny (8)} \\
mPLUG-Owl2 & LLaMA2-7B & 94.54 & 92.30 & 94.79 & 93.52 & 93.07 & 93.64 & \cellcolor{red!15} 73.05{\tiny (8)} & \cellcolor{blue!35} 73.28{\tiny (4)} & \cellcolor{red!15} 65.71{\tiny (8)} & \cellcolor{blue!10} 61.49{\tiny (9)} & \multicolumn{1}{>{\columncolor{red!10}}c|} {54.38{\tiny (9)}} & \cellcolor{blue!10} 65.58{\tiny (9)} \\
Qwen-VL-Chat & Qwen-7b & 92.33 & 91.10 & 90.06 & 91.09 & 93.19 & 91.56 & \cellcolor{red!5} 71.4{\tiny (10)} & \cellcolor{blue!5} 54.22{\tiny (10)} & \cellcolor{red!5} 63.23{\tiny (10)} & \cellcolor{blue!5} 59.79{\tiny (10)} & \multicolumn{1}{>{\columncolor{red!30}}c|} {65.09{\tiny (5)}} & \cellcolor{blue!5} 62.74{\tiny (10)} \\\hline

\multirow{2}{*}{\textbf{VLM}} & \multirow{2}{*}{\textbf{LLM}} & \multicolumn{6}{c|}{\textbf{SS} $\downarrow$} & 
\multicolumn{6}{c}{\textbf{UAcc (\%)} $\uparrow$}  \\ 
\cline{3-14} 
&  & \textbf{MMB} & \textbf{OOD} & \textbf{SQA} & \textbf{SB} & \multicolumn{1}{c|}{\textbf{AI2D}} & \textbf{Avg.} & \textbf{MMB}  & \textbf{OOD} & \textbf{SQA} & \textbf{SB} & \multicolumn{1}{c|}{\textbf{AI2D}} & \textbf{Avg.} \\ \hline 
LLaVA-v1.6-13B & Vicuna-13b & \cellcolor{red!45} 2.34{\tiny (2)} & \cellcolor{blue!35} 2.18{\tiny (4)} & \cellcolor{red!40} 2.45{\tiny (3)} & \cellcolor{blue!50} 2.49{\tiny (1)} & \multicolumn{1}{>{\columncolor{red!50}}c|} {2.33{\tiny (1)}} & \cellcolor{blue!50} 2.36{\tiny (1)} & \cellcolor{red!50} 90.41{\tiny (1)} & \cellcolor{blue!35} 86.29{\tiny (4)} & \cellcolor{red!35} 78.04{\tiny (4)} & \cellcolor{blue!50} 73.87{\tiny (1)} & \multicolumn{1}{>{\columncolor{red!50}}c|} {84.58{\tiny (1)}} & \cellcolor{blue!50} 82.64{\tiny (1)} \\
Monkey-Chat & Qwen-7b & \cellcolor{red!15} 2.7{\tiny (8)} & \cellcolor{blue!10} 2.92{\tiny (9)} & \cellcolor{red!30} 2.56{\tiny (5)} & \cellcolor{blue!10} 3.26{\tiny (9)} & \multicolumn{1}{>{\columncolor{red!10}}c|} {3.19{\tiny (9)}} & \cellcolor{blue!10} 2.93{\tiny (9)} & \cellcolor{red!25} 83.41{\tiny (6)} & \cellcolor{blue!10} 63.22{\tiny (9)} & \cellcolor{red!45} 81.7{\tiny (2)} & \cellcolor{blue!10} 52.49{\tiny (9)} & \multicolumn{1}{>{\columncolor{red!35}}c|} {56.08{\tiny (4)}} & \cellcolor{blue!20} 67.38{\tiny (7)} \\
LLaVA-v1.6-7B & Vicuna-7B & \cellcolor{red!40} 2.37{\tiny (3)} & \cellcolor{blue!25} 2.34{\tiny (6)} & \cellcolor{red!40} 2.45{\tiny (3)} & \cellcolor{blue!40} 2.53{\tiny (3)} & \multicolumn{1}{>{\columncolor{red!45}}c|} {2.37{\tiny (2)}} & \cellcolor{blue!45} 2.41{\tiny (2)} & \cellcolor{red!45} 87.87{\tiny (2)} & \cellcolor{blue!30} 82.81{\tiny (5)} & \cellcolor{red!25} 69.77{\tiny (6)} & \cellcolor{blue!45} 70.69{\tiny (2)} & \multicolumn{1}{>{\columncolor{red!45}}c|} {77.2{\tiny (2)}} & \cellcolor{blue!45} 77.67{\tiny (2)} \\
InternLM-XComposer2-VL & InternLM-7b & \cellcolor{red!5} 2.72{\tiny (10)} & \cellcolor{blue!30} 2.2{\tiny (5)} & \cellcolor{red!45} 2.41{\tiny (2)} & \cellcolor{blue!15} 3.08{\tiny (8)} & \multicolumn{1}{>{\columncolor{red!20}}c|} {3.02{\tiny (7)}} & \cellcolor{blue!20} 2.69{\tiny (7)} & \cellcolor{red!10} 69.98{\tiny (9)} & \cellcolor{blue!25} 80.49{\tiny (6)} & \cellcolor{red!50} 94.37{\tiny (1)} & \cellcolor{blue!15} 52.6{\tiny (8)} & \multicolumn{1}{>{\columncolor{red!40}}c|} {56.4{\tiny (3)}} & \cellcolor{blue!30} 70.77{\tiny (5)} \\
Yi-VL-6B & Yi-6B & \cellcolor{red!35} 2.47{\tiny (4)} & \cellcolor{blue!45} 2.02{\tiny (2)} & \cellcolor{red!10} 2.76{\tiny (9)} & \cellcolor{blue!35} 2.61{\tiny (4)} & \multicolumn{1}{>{\columncolor{red!30}}c|} {3.0{\tiny (5)}} & \cellcolor{blue!30} 2.57{\tiny (5)} & \cellcolor{red!35} 84.56{\tiny (4)} & \cellcolor{blue!45} 95.05{\tiny (2)} & \cellcolor{red!20} 64.01{\tiny (7)} & \cellcolor{blue!30} 64.56{\tiny (5)} & \multicolumn{1}{>{\columncolor{red!25}}c|} {49.31{\tiny (6)}} & \cellcolor{blue!40} 71.5{\tiny (3)} \\
CogAgent-VQA & Vicuna-7B & \cellcolor{red!50} 2.33{\tiny (1)} & \cellcolor{blue!20} 2.46{\tiny (7)} & \cellcolor{red!50} 2.36{\tiny (1)} & \cellcolor{blue!50} 2.49{\tiny (1)} & \multicolumn{1}{>{\columncolor{red!35}}c|} {2.94{\tiny (4)}} & \cellcolor{blue!40} 2.52{\tiny (3)} & \cellcolor{red!40} 85.56{\tiny (3)} & \cellcolor{blue!20} 71.14{\tiny (7)} & \cellcolor{red!30} 72.4{\tiny (5)} & \cellcolor{blue!40} 69.96{\tiny (3)} & \multicolumn{1}{>{\columncolor{red!20}}c|} {49.0{\tiny (7)}} & \cellcolor{blue!25} 69.61{\tiny (6)} \\
MobileVLMV2-7B & Vicuna-7B & \cellcolor{red!30} 2.53{\tiny (5)} & \cellcolor{blue!15} 2.61{\tiny (8)} & \cellcolor{red!25} 2.62{\tiny (6)} & \cellcolor{blue!25} 2.8{\tiny (6)} & \multicolumn{1}{>{\columncolor{red!5}}c|} {3.4{\tiny (10)}} & \cellcolor{blue!15} 2.79{\tiny (8)} & \cellcolor{red!30} 84.19{\tiny (5)} & \cellcolor{blue!15} 64.35{\tiny (8)} & \cellcolor{red!40} 79.07{\tiny (3)} & \cellcolor{blue!25} 62.18{\tiny (6)} & \multicolumn{1}{>{\columncolor{red!5}}c|} {39.39{\tiny (10)}} & \cellcolor{blue!15} 65.84{\tiny (8)} \\
MoE-LLaVA-Phi2-2.7B & Phi2-2.7B & \cellcolor{red!25} 2.54{\tiny (6)} & \cellcolor{blue!50} 1.89{\tiny (1)} & \cellcolor{red!20} 2.7{\tiny (7)} & \cellcolor{blue!30} 2.69{\tiny (5)} & \multicolumn{1}{>{\columncolor{red!40}}c|} {2.92{\tiny (3)}} & \cellcolor{blue!35} 2.55{\tiny (4)} & \cellcolor{red!20} 82.83{\tiny (7)} & \cellcolor{blue!50} 100.73{\tiny (1)} & \cellcolor{red!10} 61.18{\tiny (9)} & \cellcolor{blue!35} 64.67{\tiny (4)} & \multicolumn{1}{>{\columncolor{red!15}}c|} {47.87{\tiny (8)}} & \cellcolor{blue!35} 71.46{\tiny (4)} \\
mPLUG-Owl2 & LLaMA2-7B & \cellcolor{red!20} 2.55{\tiny (7)} & \cellcolor{blue!40} 2.09{\tiny (3)} & \cellcolor{red!15} 2.71{\tiny (8)} & \cellcolor{blue!20} 2.93{\tiny (7)} & \multicolumn{1}{>{\columncolor{red!30}}c|} {3.0{\tiny (5)}} & \cellcolor{blue!25} 2.65{\tiny (6)} & \cellcolor{red!15} 78.4{\tiny (8)} & \cellcolor{blue!40} 89.24{\tiny (3)} & \cellcolor{red!15} 62.92{\tiny (8)} & \cellcolor{blue!20} 52.91{\tiny (7)} & \multicolumn{1}{>{\columncolor{red!10}}c|} {45.09{\tiny (9)}} & \cellcolor{blue!10} 65.71{\tiny (9)} \\
Qwen-VL-Chat & Qwen-7b & \cellcolor{red!15} 2.7{\tiny (8)} & \cellcolor{blue!5} 3.32{\tiny (10)} & \cellcolor{red!5} 2.9{\tiny (10)} & \cellcolor{blue!5} 3.32{\tiny (10)} & \multicolumn{1}{>{\columncolor{red!15}}c|} {3.1{\tiny (8)}} & \cellcolor{blue!5} 3.07{\tiny (10)} & \cellcolor{red!5} 69.58{\tiny (10)} & \cellcolor{blue!5} 40.28{\tiny (10)} & \cellcolor{red!5} 54.71{\tiny (10)} & \cellcolor{blue!5} 44.7{\tiny (10)} & \multicolumn{1}{>{\columncolor{red!30}}c|} {54.3{\tiny (5)}} & \cellcolor{blue!5} 52.71{\tiny (10)} \\ \hline

\end{tabular}

}
\vspace*{0.15cm}
\caption{Results of VLMs with LLM sizes ranging from 2.7B to 13B. These results represent the mean values of LAC and APS. Each row contains a relative ranking among all models in the corresponding column (a small number in parentheses), and this ranking is also highlighted with color.}
\label{tab:main}
\end{table*}

\section{Experiment Setup}

\subsection{Evaluation Metrics}

We follow the evaluation approach from \cite{ye2024benchmarking}. We also asses Vision-Language Models by considering two aspects: prediction accuracy and prediction uncertainty. To calculate the accuracy, we utilize the logits generated for the last prompt token and select six corresponding tokens associated with letters from 'A' to 'F.' We consider that the model answers with the letter associated with the maximum logit value.

To evaluate model uncertainty, we also calculate prediction sets for every instance in the test set using two score functions: LAC and APS. We first average lengths of the prediction sets over the test dataset and then average by score functions to get final metric Set Sizes (SS).

The last metric from \cite{ye2024benchmarking}, Uncertainty-aware Accuracy (UAcc), combines both accuracy and uncertainty:
\begin{equation}
    UAcc = \frac{Acc}{SS}\sqrt{|Y|}
\end{equation}
where $Y$ denotes the option set. Because of such a definition, UAcc takes values in the range $[0, \sqrt{|\mathcal{Y}|}]$, thus can be larger than 1.

Besides, we also calculate two naive model calibration metrics - Expected Calibration Error  (ECE) and Maximum Calibration Error (MCE) to check how much it is correlated with accuracy and conformal prediction. ECE measures the average discrepancy between the confidence if the model's predictions and their accuracy and is defined as:

\begin{equation}
    ECE = \sum_{m=1}^{M} \frac{|B_m|}{n} | acc(B_m) - conf(B_m) |
\end{equation}
where
\begin{equation}
    acc(B_m) = \frac{1}{|B_m|} \sum_{i \in B_m} \mathbbm{1}(\hat{y}_i = y_i)
\end{equation}
\begin{equation}
    conf(B_m) = \frac{1}{B_m} \sum_{i \in B_m} \hat{p}_i
\end{equation}
$\hat{y}_i, y_i$ are the predicted and true class labels for sample $i$,  $B_m$ is the set of indices of samples whose prediction confidence falls, $\hat{p}_i$ is the confidence for sample $i$.  

\begin{figure*}[h]
  \centering
  \includegraphics[width=\linewidth]{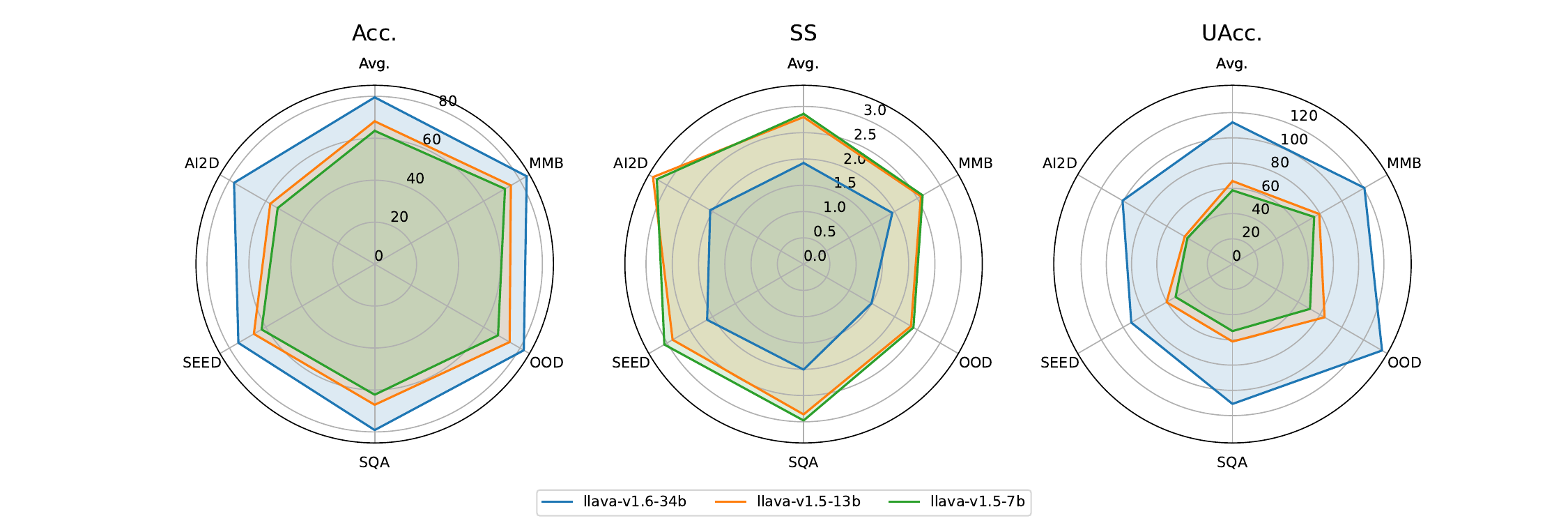}
  \caption{Comparison of LLaVA1.6 with LLM of different size.}
  \label{fig:scale_llava1.6}
\end{figure*}

\subsection{Prompting Strategies}

To construct a prompt for VLM, we adapt the prompt template for multiple-choice VQA tasks evaluation from  LLaVA \cite{LLAVA}. At the beginning of the prompt, we include the text of the question itself, as well as a hint if present in the question. Further, we add 6 answer options, one per line. Before each answer option, we indicate the corresponding letter. Following this, on a new line, we also add the standard phrase "Answer with the option's letter from the given choices directly." The resulting text is wrapped in the standard prompt template for each model.

More specifically, for the LLaVA, Yi-VL, Qwen, Monkey, MoE-LLaVA, mPLUG-Owl, MobileVLM models, we utilize prompt templates from the respective Github repositories. For the CogAgent and InternLM-XComposer2 models, we use prompt templates from their Hugging Face repositories.

We do not include any demonstrations in the prompt, as many models are only capable of supporting a single image in their input.

\subsection{Implementation Details}

In our experiments, we follow the setup from \cite{ye2024benchmarking} and utilize LAC and APS score functions with error rate $\alpha$ equal to 0.1, which means we aimed for a 90\% probability of including the true label in the prediction set. 

For each task, we divided the dataset into a calibration set (50\% of the data) and a test set (the remaining 50\%). The reported results are the average values obtained from two conformal score functions. Both scores fulfill the coverage guarantee requirement but may produce different sizes of prediction sets. By averaging the score functions, we aimed to ensure a more rigorous evaluation of uncertainty and mitigate the influence of different scoring methods on the overall assessment.

All reported results are based on the test set, aiming for a comprehensive and reliable analysis of the models' performance.

\begin{figure*}[h]
  \centering
  \includegraphics[width=\linewidth]{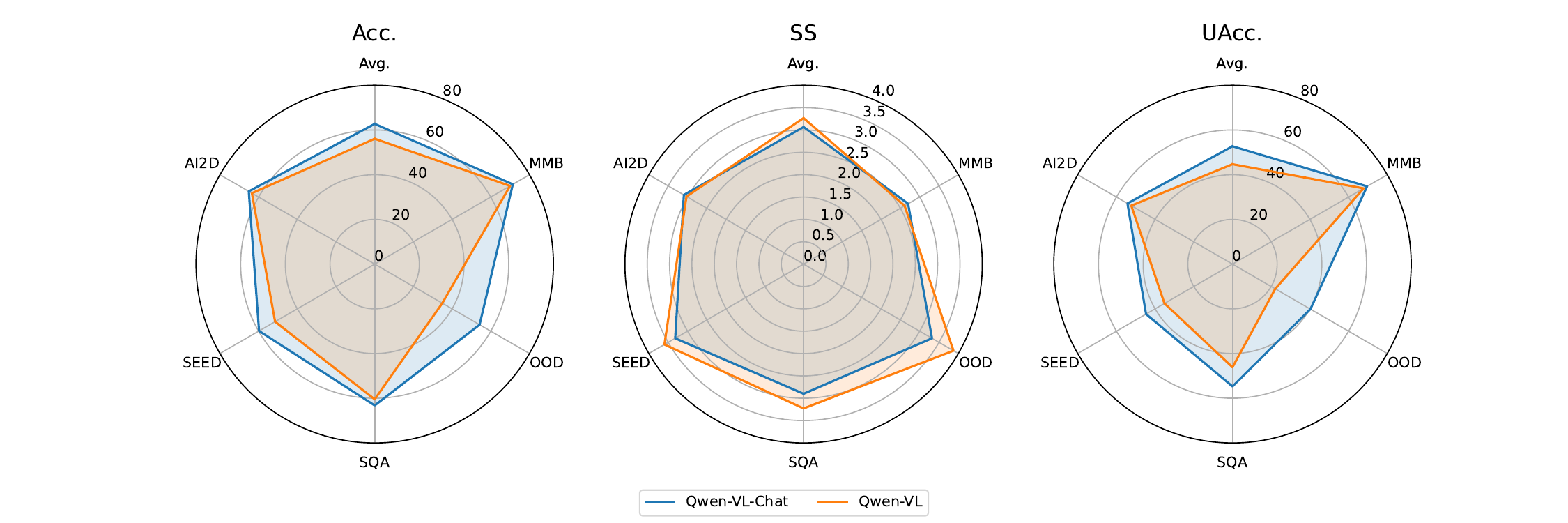}
  \caption{Comparison of Qwen-VL and Qwen-VL-Chat.}
  \label{fig:base_vs_chat_qwen}
\end{figure*}

\section{Evaluation and Analysis}

\subsection{Comparative Analyses}

The main results of our study are presented in Table ~\ref{tab:main}. Similar to findings in ~\cite{ye2024benchmarking} Vision-Language Models in our benchmark surpasses the 90\% threshold on coverage rate, and each model managed to achieve this on all datasets. This indicates the reliability of choosing prediction sets for uncertainty assessment both for LLMs and VLMs.

Accuracy and uncertainty for VLMs in our experiments are not aligned that is increase in accuracy increasing does not entail decrease in uncertainty. It can be observed by the relative ranking of several models. Monkey-Chat stands out especially: being second-ranked by accuracy, it only ranked ninth by Set Sizes. Moreover, it never ranked higher than fifth and ranked eighth and below on all benchmarks except ScienceQA. There are other examples of models whose relative rankings differ significantly for accuracy and uncertainty: MoE-LLaVA-Phi2-2.7B ranked eighth in accuracy while being fourth on Set Sizes. 

Such behavior results in significant difference of model's ranking on Acc and UAcc metrics. Due to high uncertainty, the UAcc value for the Monkey-Chat model becomes lower than accuracy, and the model drops from second to seventh place in our rating. On the contrary, due to the low uncertainty, MoE-LLaVA-Phi2-2.7B has higher UAcc and rises from eighth to fourth rank. 

Even with a significant difference in accuracy, there may be a situation where an outperforming model has higher uncertainty. For example, Monkey-Chat outperforms MoE-LLaVA-Phi2-2.7B in accuracy by 12.19 absolute points on the AI2D dataset yet shows higher Set Sizes value. Similar cases are observed for other pairs of models: InternLM-XComposer2-VL outperforms MoE-LLaVA-Phi2-2.7B on the AI2D dataset by 10.37 and demonstrates higher uncertainty. 

Only one case can be noticed in Table ~\ref{tab:main} where the value of UAcc exceeded 100\%. However, among the models examined, similar cases occurred several more times, particularly in larger models. The Yi-VL-34B model exceeded this value on the MMBench and OODCV datasets, while the LLaVA-v1.6-34B model exceeded this value on all datasets except SEEDBench and also surpassed this value on average.

It is worth noting the importance of including OODCV and AI2D datasets in our benchmark, as they identify the largest ranges of values for both accuracy and uncertainty. On these datasets, we observe an accuracy range from 54.22 to 74.82 and from 53.55 to 73.67, respectively, while on the other datasets, these ranges are much smaller. Similarly, Set Sizes range from 1.89 to 3.32 on OODCV and from 2.33 to 3.4 on AI2D, while on other datasets, these ranges are smaller than 1.

Finally, as can be seen from Table ~\ref{tab:ef}, presented in Appendix \ref{app:ef}, an anomalously large E and F rate is observed specifically in the OOD dataset among certain models.

\begin{table*}[h]
\centering
\resizebox{\textwidth}{!}{%
\begin{tabular}{l|l|cccccc|cccccc}
\hline

\multirow{2}{*}{\textbf{VLM}} & \multirow{2}{*}{\textbf{LLM}} & \multicolumn{6}{c|}{\textbf{ECE} $\downarrow$} & 
\multicolumn{6}{c}{\textbf{MCE (\%)} $\downarrow$}  \\ 
\cline{3-14} 
&  & \textbf{MMB} & \textbf{OOD} & \textbf{SQA} & \textbf{SB} & \multicolumn{1}{c|}{\textbf{AI2D}} & \textbf{Avg.} & \textbf{MMB}  & \textbf{OOD} & \textbf{SQA} & \textbf{SB} & \multicolumn{1}{c|}{\textbf{AI2D}} & \textbf{Avg.} \\ \hline 

LLaVA-v1.6-13B & Vicuna-13b & \cellcolor{red!20} 6.12{\tiny (7)} & \cellcolor{blue!50} 2.22{\tiny (1)} & \cellcolor{red!15} 9.88{\tiny (8)} & \cellcolor{blue!20} 11.53{\tiny (7)} & \multicolumn{1}{>{\columncolor{red!45}}c|} {2.68{\tiny (2)}} & \cellcolor{blue!30} 6.49{\tiny (5)} & \cellcolor{red!25} 18.35{\tiny (6)} & \cellcolor{blue!15} 25.59{\tiny (8)} & \cellcolor{red!15} 29.54{\tiny (8)} & \cellcolor{blue!30} 20.82{\tiny (5)} & \multicolumn{1}{>{\columncolor{red!40}}c|} {10.73{\tiny (3)}} & \cellcolor{blue!25} 21.0{\tiny (6)} \\
Monkey-Chat & Qwen-7b & \cellcolor{red!40} 2.44{\tiny (3)} & \cellcolor{blue!5} 11.85{\tiny (10)} & \cellcolor{red!45} 2.91{\tiny (2)} & \cellcolor{blue!40} 5.08{\tiny (3)} & \multicolumn{1}{>{\columncolor{red!35}}c|} {3.21{\tiny (4)}} & \cellcolor{blue!40} 5.1{\tiny (3)} & \cellcolor{red!45} 7.95{\tiny (2)} & \cellcolor{blue!30} 19.73{\tiny (5)} & \cellcolor{red!10} 30.99{\tiny (9)} & \cellcolor{blue!10} 26.87{\tiny (9)} & \multicolumn{1}{>{\columncolor{red!30}}c|} {13.01{\tiny (5)}} & \cellcolor{blue!35} 19.71{\tiny (4)} \\
LLaVA-v1.6-7B & Vicuna-7B & \cellcolor{red!10} 7.52{\tiny (9)} & \cellcolor{blue!35} 3.72{\tiny (4)} & \cellcolor{red!10} 10.23{\tiny (9)} & \cellcolor{blue!25} 11.44{\tiny (6)} & \multicolumn{1}{>{\columncolor{red!50}}c|} {2.63{\tiny (1)}} & \cellcolor{blue!25} 7.11{\tiny (6)} & \cellcolor{red!10} 24.86{\tiny (9)} & \cellcolor{blue!5} 74.4{\tiny (10)} & \cellcolor{red!25} 24.25{\tiny (6)} & \cellcolor{blue!25} 20.96{\tiny (6)} & \multicolumn{1}{>{\columncolor{red!50}}c|} {7.57{\tiny (1)}} & \cellcolor{blue!5} 30.41{\tiny (10)} \\
InternLM-XComposer2-VL & InternLM-7b & \cellcolor{red!25} 5.95{\tiny (6)} & \cellcolor{blue!45} 3.46{\tiny (2)} & \cellcolor{red!40} 3.78{\tiny (3)} & \cellcolor{blue!30} 10.17{\tiny (5)} & \multicolumn{1}{>{\columncolor{red!25}}c|} {6.24{\tiny (6)}} & \cellcolor{blue!35} 5.92{\tiny (4)} & \cellcolor{red!20} 19.69{\tiny (7)} & \cellcolor{blue!40} 14.8{\tiny (3)} & \cellcolor{red!5} 34.56{\tiny (10)} & \cellcolor{blue!40} 15.99{\tiny (3)} & \multicolumn{1}{>{\columncolor{red!25}}c|} {14.56{\tiny (6)}} & \cellcolor{blue!30} 19.92{\tiny (5)} \\
Yi-VL-6B & Yi-6B & \cellcolor{red!30} 5.6{\tiny (5)} & \cellcolor{blue!25} 7.72{\tiny (6)} & \cellcolor{red!25} 7.64{\tiny (6)} & \cellcolor{blue!15} 13.37{\tiny (8)} & \multicolumn{1}{>{\columncolor{red!15}}c|} {13.27{\tiny (8)}} & \cellcolor{blue!15} 9.52{\tiny (8)} & \cellcolor{red!35} 17.51{\tiny (4)} & \cellcolor{blue!10} 26.28{\tiny (9)} & \cellcolor{red!30} 19.54{\tiny (5)} & \cellcolor{blue!20} 23.46{\tiny (7)} & \multicolumn{1}{>{\columncolor{red!15}}c|} {24.37{\tiny (8)}} & \cellcolor{blue!20} 22.23{\tiny (7)} \\
CogAgent-VQA & Vicuna-7B & \cellcolor{red!50} 1.53{\tiny (1)} & \cellcolor{blue!10} 10.98{\tiny (9)} & \cellcolor{red!50} 2.88{\tiny (1)} & \cellcolor{blue!45} 4.84{\tiny (2)} & \multicolumn{1}{>{\columncolor{red!30}}c|} {4.33{\tiny (5)}} & \cellcolor{blue!45} 4.91{\tiny (2)} & \cellcolor{red!50} 5.42{\tiny (1)} & \cellcolor{blue!25} 22.91{\tiny (6)} & \cellcolor{red!40} 15.63{\tiny (3)} & \cellcolor{blue!45} 11.89{\tiny (2)} & \multicolumn{1}{>{\columncolor{red!45}}c|} {9.85{\tiny (2)}} & \cellcolor{blue!45} 13.14{\tiny (2)} \\
MobileVLMV2-7B & Vicuna-7B & \cellcolor{red!15} 7.13{\tiny (8)} & \cellcolor{blue!15} 10.07{\tiny (8)} & \cellcolor{red!20} 7.68{\tiny (7)} & \cellcolor{blue!10} 14.91{\tiny (9)} & \multicolumn{1}{>{\columncolor{red!10}}c|} {15.44{\tiny (9)}} & \cellcolor{blue!10} 11.04{\tiny (9)} & \cellcolor{red!15} 20.28{\tiny (8)} & \cellcolor{blue!20} 25.07{\tiny (7)} & \cellcolor{red!35} 19.41{\tiny (4)} & \cellcolor{blue!15} 25.67{\tiny (8)} & \multicolumn{1}{>{\columncolor{red!10}}c|} {26.43{\tiny (9)}} & \cellcolor{blue!15} 23.37{\tiny (8)} \\
MoE-LLaVA-Phi2-2.7B & Phi2-2.7B & \cellcolor{red!35} 4.28{\tiny (4)} & \cellcolor{blue!20} 9.52{\tiny (7)} & \cellcolor{red!30} 6.83{\tiny (5)} & \cellcolor{blue!35} 7.35{\tiny (4)} & \multicolumn{1}{>{\columncolor{red!20}}c|} {8.51{\tiny (7)}} & \cellcolor{blue!20} 7.3{\tiny (7)} & \cellcolor{red!30} 17.87{\tiny (5)} & \cellcolor{blue!35} 16.55{\tiny (4)} & \cellcolor{red!45} 14.66{\tiny (2)} & \cellcolor{blue!35} 19.22{\tiny (4)} & \multicolumn{1}{>{\columncolor{red!20}}c|} {19.35{\tiny (7)}} & \cellcolor{blue!40} 17.53{\tiny (3)} \\
mPLUG-Owl2 & LLaMA2-7B & \cellcolor{red!5} 10.65{\tiny (10)} & \cellcolor{blue!30} 5.39{\tiny (5)} & \cellcolor{red!5} 10.69{\tiny (10)} & \cellcolor{blue!5} 17.9{\tiny (10)} & \multicolumn{1}{>{\columncolor{red!5}}c|} {17.12{\tiny (10)}} & \cellcolor{blue!5} 12.35{\tiny (10)} & \cellcolor{red!5} 27.03{\tiny (10)} & \cellcolor{blue!45} 11.27{\tiny (2)} & \cellcolor{red!20} 24.39{\tiny (7)} & \cellcolor{blue!5} 32.82{\tiny (10)} & \multicolumn{1}{>{\columncolor{red!5}}c|} {30.56{\tiny (10)}} & \cellcolor{blue!10} 25.21{\tiny (9)} \\
Qwen-VL-Chat & Qwen-7b & \cellcolor{red!45} 1.88{\tiny (2)} & \cellcolor{blue!40} 3.7{\tiny (3)} & \cellcolor{red!35} 5.0{\tiny (4)} & \cellcolor{blue!50} 3.95{\tiny (1)} & \multicolumn{1}{>{\columncolor{red!40}}c|} {3.1{\tiny (3)}} & \cellcolor{blue!50} 3.52{\tiny (1)} & \cellcolor{red!40} 9.18{\tiny (3)} & \cellcolor{blue!50} 10.03{\tiny (1)} & \cellcolor{red!50} 12.08{\tiny (1)} & \cellcolor{blue!50} 7.49{\tiny (1)} & \multicolumn{1}{>{\columncolor{red!35}}c|} {11.86{\tiny (4)}} & \cellcolor{blue!50} 10.13{\tiny (1)} \\ \hline

\end{tabular}

}
\vspace*{0.15cm}

\caption{ECE and MCE values of VLMs with LLM sizes ranging from 2.7B to 13B. Each row contains a relative ranking among all models in the corresponding column (a small number in parentheses), and this ranking is also highlighted with color.}
\label{tab:ece}
\end{table*}

\subsection{Effects of LLM Scale}

Increasing the size of LLMs was observed to improve accuracy, as illustrated in the case of LLaVA1.6 (see Figure \ref{fig:scale_llava1.6}). When the size of LLM grows from 7 to 13B, the model accuracy increases slightly. However, when the size of LLM grows from 13 to 34B, the gap in accuracy becomes very noticeable. Similar behavior was observed in terms of Set Sizes. While LLaVA-v1.6-7B and LLaVA-v1.6-13B show comparable set sizes, LLaVA-v1.6-34B demonstrates much lower uncertainty.

Additional comparisons are provided in Appendix \ref{app:scale}. For the Yi-VL model series, we also observe an increase in accuracy and a decrease in uncertainty when increasing the language model from 6B to 34B.

In contrast, for MobileVLM, gaps in accuracy are noted, but uncertainty values are comparable.
Moreover, across five datasets, each model from the MobileVLMV2 series could have either the lowest or highest set size value.

Thus, it can be suggested that uncertainty decreases only with a very significant increase in the language model in VLMs.

\subsection{Effects of Chat Finetuning}

In most cases, we observed that models finetuned for chat versions outperformed the base models in terms of accuracy. However, this is different when we compare models regarding the set sizes. While in most cases, this holds true as well, we observed situations where the uncertainty of the base model was lower than that of the chat finetuned version.

As shown in Figure ~\ref{fig:base_vs_chat_qwen}, Qwen-VL and Qwen-VL-Chat demonstrate comparable set sizes across all benchmarks except OODCV. There is also a significant difference in accuracy in the OODCV benchmark - the chat version remarkably outperformed the non-chat model.
It is noticeable that accuracy and Uacc values are comparable on SQA and MMB benchmarks, and Qwen-VL-Chat shows improvement in the OODCV benchmark.

Additional results on the InternLM-XComposer2 and Monkey models are presented in Appendix ~\ref{sec:base_vs_chat}.

We also attempted to conduct an experiment with the CogVLM-base-224 and CogVLM-base-490 models to compare them with CogVLM-Chat. However, we were unable to get responses to multiple-choice questions in the form of a single letter from them.

\subsection{Calibration Error Results}

The ECE and MCE metrics results are provided in Table \ref{tab:ece}. Similar to previous findings ~\cite{kadavath2022language, kumar2023conformal}, we observe that all models are well-calibrated across datasets across all five datasets, and their mean ECE values are reasonably low. The best model in terms of ECE is Qwen-VL-Chat, with a mean value of 3.07\% across all datasets, and mPLUG-Owl2 demonstrates the highest ECE across all models with a value of 12.35\%. Moreover, we do not observe significant variation between different datasets, with the lowest mean across models equal to 5.31\% on MMBench and the highest equal to 10.05\% on SEEDBench. We also observe that the results based on the ECE metric are not correlated with the results of conformal prediction. Moreover, the last ranking model in terms of Set Sizes, Qwen-VL-Chat, is the best in terms of ECE. Besides being the best model in terms of conformal prediction, LLaVA-v1.6-13B ranks only fifth in the ECE metric.

The result on Vision-Language Models MCE also aligns with corresponding results on LLM ~\cite{kumar2023conformal}. All assessed models have quite high mean MCE values ranging from 10.13\% on Qwen-VL-Chat to 30.41\% on LLaVA-v1.6-7B. One can also observe a huge MCE equal to 74.4\% demonstrated by LLaVA-v1.6-7B on OODCV.

These findings suggest that utilizing these metrics is not an appropriate choice for uncertainty estimation due to their lack of formal guarantees and the absence of correlation with the above metrics. 

\subsection{Other Findings}
As in ~\cite{ye2024benchmarking}, we check how much the variation of the calibration set size impacts Coverage Rate and Set Sizes. We don't observe any significant changes in the values of these metrics when adjusting the size of the calibration set from 10 to 50\% of the total dataset size.
Thus, our results for VLMs are in agreement with similar results for LLMs.

We also perform a detailed analysis of SEEDBench and OODCV results by their internal categories. These results can be found in Appendices ~\ref{app:seed}, ~\ref{app:oodcv}.

\section{Conclusion}

In this study, we comprehensively compared Vision-Language Models in terms of their accuracy uncertainty. We identified situations where accuracy and uncertainty may not align, emphasizing the insufficiency of relying solely on accuracy when assessing models. Moreover, situations have been identified where these two metrics contradict each other in the relative comparison of models. This highlights the importance of evaluating models across multiple benchmarks that consider various metrics in order to create responsible and trusted artificial intelligence. 

\section*{Limitations}

In this study, we only investigate multiple-choice VQA tasks, which cover only a portion of the benchmarks used to evaluate multimodal models. In addition, we examine only the standard VQA prompting strategy from LLaVA. Since the majority of the inspected VLMs do not support multi-image input, we do not conduct experiments with demonstrations in prompts. However, this could help establish new patterns in VLMs uncertainty. 

A significant direction for future research on multimodal models could also involve assessing their uncertainty in other Vision-Language tasks: open-ended VQA, Image Captioning, and Visual Grounding.

\twocolumn
\clearpage






%
\bibliographystyle{IEEEtran}

\clearpage

\appendices

\begin{table*}[h]
\centering
\resizebox{\textwidth}{!}{%
\begin{tabular}{l|l|cccccc|cccccc}
\hline
\multirow{2}{*}{\textbf{VLM}} & \multirow{2}{*}{\textbf{LLM}} & \multicolumn{6}{c|}{\textbf{Coverage Rate (\%)}} & 
\multicolumn{6}{c}{\textbf{Acc (\%)} $\uparrow$} \\ 
\cline{3-14} 
&  & \textbf{MMB} & \textbf{OOD} & \textbf{SQA} & \textbf{SB} & \multicolumn{1}{c|}{\textbf{AI2D}} & \textbf{Avg.} & \textbf{MMB}  & \textbf{OOD} & \textbf{SQA} & \textbf{SB} & \multicolumn{1}{c|}{\textbf{AI2D}} & \textbf{Avg.} \\ \hline 
LLaVA-v1.6-13B & Vicuna-13b & 90.18 & 91.00 & 89.28 & 89.84 & 90.47 & 90.15 & \cellcolor{red!45} 76.75{\tiny (2)} & \cellcolor{blue!30} 72.93{\tiny (5)} & \cellcolor{red!35} 70.56{\tiny (4)} & \cellcolor{blue!50} 70.37{\tiny (1)} & \multicolumn{1}{>{\columncolor{red!50}}c|} {73.67{\tiny (1)}} & \cellcolor{blue!50} 72.85{\tiny (1)} \\
Monkey-Chat & Qwen-7b & 89.45 & 88.75 & 90.44 & 89.22 & 90.98 & 89.77 & \cellcolor{red!50} 76.98{\tiny (1)} & \cellcolor{blue!25} 70.6{\tiny (6)} & \cellcolor{red!45} 74.66{\tiny (2)} & \cellcolor{blue!20} 66.1{\tiny (7)} & \multicolumn{1}{>{\columncolor{red!40}}c|} {67.95{\tiny (3)}} & \cellcolor{blue!45} 71.26{\tiny (2)} \\
LLaVA-v1.6-7B & Vicuna-7B & 89.26 & 89.10 & 89.83 & 90.19 & 89.65 & 89.61 & \cellcolor{red!35} 75.56{\tiny (4)} & \cellcolor{blue!40} 73.7{\tiny (3)} & \cellcolor{red!20} 65.86{\tiny (7)} & \cellcolor{blue!45} 69.06{\tiny (2)} & \multicolumn{1}{>{\columncolor{red!45}}c|} {69.75{\tiny (2)}} & \cellcolor{blue!40} 70.78{\tiny (3)} \\
InternLM-XComposer2-VL & InternLM-7b & 89.17 & 88.96 & 89.58 & 89.90 & 89.87 & 89.50 & \cellcolor{red!10} 71.77{\tiny (9)} & \cellcolor{blue!20} 70.04{\tiny (7)} & \cellcolor{red!50} 77.95{\tiny (1)} & \cellcolor{blue!15} 64.44{\tiny (8)} & \multicolumn{1}{>{\columncolor{red!35}}c|} {66.13{\tiny (4)}} & \cellcolor{blue!35} 70.07{\tiny (4)} \\
Yi-VL-6B & Yi-6B & 90.22 & 89.94 & 89.78 & 89.84 & 91.01 & 90.16 & \cellcolor{red!30} 75.24{\tiny (5)} & \cellcolor{blue!45} 73.91{\tiny (2)} & \cellcolor{red!25} 66.72{\tiny (6)} & \cellcolor{blue!25} 66.25{\tiny (6)} & \multicolumn{1}{>{\columncolor{red!25}}c|} {58.84{\tiny (6)}} & \cellcolor{blue!30} 68.19{\tiny (5)} \\
CogAgent-VQA & Vicuna-7B & 90.68 & 90.37 & 90.14 & 89.36 & 90.65 & 90.24 & \cellcolor{red!25} 74.78{\tiny (6)} & \cellcolor{blue!15} 68.57{\tiny (8)} & \cellcolor{red!30} 67.12{\tiny (5)} & \cellcolor{blue!40} 68.01{\tiny (3)} & \multicolumn{1}{>{\columncolor{red!20}}c|} {58.2{\tiny (7)}} & \cellcolor{blue!25} 67.34{\tiny (6)} \\
MobileVLMV2-7B & Vicuna-7B & 89.63 & 90.86 & 89.07 & 89.49 & 90.23 & 89.86 & \cellcolor{red!40} 75.97{\tiny (3)} & \cellcolor{blue!10} 66.53{\tiny (9)} & \cellcolor{red!40} 72.33{\tiny (3)} & \cellcolor{blue!35} 66.71{\tiny (4)} & \multicolumn{1}{>{\columncolor{red!5}}c|} {53.55{\tiny (10)}} & \cellcolor{blue!20} 67.02{\tiny (7)} \\
MoE-LLaVA-Phi2-2.7B & Phi2-2.7B & 89.26 & 89.17 & 90.84 & 89.66 & 90.08 & 89.80 & \cellcolor{red!20} 73.73{\tiny (7)} & \cellcolor{blue!50} 74.82{\tiny (1)} & \cellcolor{red!10} 64.04{\tiny (9)} & \cellcolor{blue!30} 66.42{\tiny (5)} & \multicolumn{1}{>{\columncolor{red!15}}c|} {55.76{\tiny (8)}} & \cellcolor{blue!15} 66.95{\tiny (8)} \\
mPLUG-Owl2 & LLaMA2-7B & 89.81 & 89.52 & 91.40 & 89.94 & 90.34 & 90.20 & \cellcolor{red!15} 73.05{\tiny (8)} & \cellcolor{blue!35} 73.28{\tiny (4)} & \cellcolor{red!15} 65.71{\tiny (8)} & \cellcolor{blue!10} 61.49{\tiny (9)} & \multicolumn{1}{>{\columncolor{red!10}}c|} {54.38{\tiny (9)}} & \cellcolor{blue!10} 65.58{\tiny (9)} \\
Qwen-VL-Chat & Qwen-7b & 88.44 & 88.75 & 88.11 & 89.21 & 89.69 & 88.84 & \cellcolor{red!5} 71.4{\tiny (10)} & \cellcolor{blue!5} 54.22{\tiny (10)} & \cellcolor{red!5} 63.23{\tiny (10)} & \cellcolor{blue!5} 59.79{\tiny (10)} & \multicolumn{1}{>{\columncolor{red!30}}c|} {65.09{\tiny (5)}} & \cellcolor{blue!5} 62.74{\tiny (10)} \\ \hline

\multirow{2}{*}{\textbf{VLM}} & \multirow{2}{*}{\textbf{LLM}} & \multicolumn{6}{c|}{\textbf{SS} $\downarrow$} & 
\multicolumn{6}{c}{\textbf{UAcc (\%)} $\uparrow$}  \\ 
\cline{3-14} 
&  & \textbf{MMB} & \textbf{OOD} & \textbf{SQA} & \textbf{SB} & \multicolumn{1}{c|}{\textbf{AI2D}} & \textbf{Avg.} & \textbf{MMB}  & \textbf{OOD} & \textbf{SQA} & \textbf{SB} & \multicolumn{1}{c|}{\textbf{AI2D}} & \textbf{Avg.} \\ \hline 
LLaVA-v1.6-13B & Vicuna-13b & \cellcolor{red!50} 1.56{\tiny (1)} & \cellcolor{blue!40} 1.68{\tiny (3)} & \cellcolor{red!35} 1.69{\tiny (4)} & \cellcolor{blue!50} 1.86{\tiny (1)} & \multicolumn{1}{>{\columncolor{red!50}}c|} {1.65{\tiny (1)}} & \cellcolor{blue!50} 1.69{\tiny (1)} & \cellcolor{red!50} 120.71{\tiny (1)} & \cellcolor{blue!35} 106.06{\tiny (4)} & \cellcolor{red!35} 102.37{\tiny (4)} & \cellcolor{blue!50} 92.64{\tiny (1)} & \multicolumn{1}{>{\columncolor{red!50}}c|} {109.33{\tiny (1)}} & \cellcolor{blue!50} 106.22{\tiny (1)} \\
Monkey-Chat & Qwen-7b & \cellcolor{red!35} 1.61{\tiny (4)} & \cellcolor{blue!10} 2.18{\tiny (9)} & \cellcolor{red!40} 1.66{\tiny (3)} & \cellcolor{blue!15} 2.5{\tiny (8)} & \multicolumn{1}{>{\columncolor{red!40}}c|} {2.35{\tiny (3)}} & \cellcolor{blue!20} 2.06{\tiny (7)} & \cellcolor{red!40} 117.02{\tiny (3)} & \cellcolor{blue!15} 79.31{\tiny (8)} & \cellcolor{red!45} 110.46{\tiny (2)} & \cellcolor{blue!20} 64.63{\tiny (7)} & \multicolumn{1}{>{\columncolor{red!40}}c|} {70.95{\tiny (3)}} & \cellcolor{blue!25} 88.48{\tiny (6)} \\
LLaVA-v1.6-7B & Vicuna-7B & \cellcolor{red!45} 1.58{\tiny (2)} & \cellcolor{blue!30} 1.73{\tiny (5)} & \cellcolor{red!30} 1.87{\tiny (5)} & \cellcolor{blue!45} 1.94{\tiny (2)} & \multicolumn{1}{>{\columncolor{red!45}}c|} {1.76{\tiny (2)}} & \cellcolor{blue!45} 1.78{\tiny (2)} & \cellcolor{red!45} 117.06{\tiny (2)} & \cellcolor{blue!30} 104.65{\tiny (5)} & \cellcolor{red!25} 86.31{\tiny (6)} & \cellcolor{blue!45} 87.0{\tiny (2)} & \multicolumn{1}{>{\columncolor{red!45}}c|} {96.98{\tiny (2)}} & \cellcolor{blue!45} 98.4{\tiny (2)} \\
InternLM-XComposer2-VL & InternLM-7b & \cellcolor{red!10} 1.97{\tiny (9)} & \cellcolor{blue!25} 1.82{\tiny (6)} & \cellcolor{red!50} 1.44{\tiny (1)} & \cellcolor{blue!10} 2.58{\tiny (9)} & \multicolumn{1}{>{\columncolor{red!35}}c|} {2.36{\tiny (4)}} & \cellcolor{blue!30} 2.03{\tiny (5)} & \cellcolor{red!10} 89.43{\tiny (9)} & \cellcolor{blue!25} 94.34{\tiny (6)} & \cellcolor{red!50} 132.31{\tiny (1)} & \cellcolor{blue!10} 61.09{\tiny (9)} & \multicolumn{1}{>{\columncolor{red!35}}c|} {68.7{\tiny (4)}} & \cellcolor{blue!35} 89.17{\tiny (4)} \\
Yi-VL-6B & Yi-6B & \cellcolor{red!30} 1.62{\tiny (5)} & \cellcolor{blue!45} 1.54{\tiny (2)} & \cellcolor{red!20} 2.01{\tiny (7)} & \cellcolor{blue!30} 2.11{\tiny (5)} & \multicolumn{1}{>{\columncolor{red!20}}c|} {2.51{\tiny (7)}} & \cellcolor{blue!35} 1.96{\tiny (4)} & \cellcolor{red!25} 113.71{\tiny (6)} & \cellcolor{blue!45} 117.88{\tiny (2)} & \cellcolor{red!20} 81.36{\tiny (7)} & \cellcolor{blue!25} 77.05{\tiny (6)} & \multicolumn{1}{>{\columncolor{red!25}}c|} {57.33{\tiny (6)}} & \cellcolor{blue!40} 89.46{\tiny (3)} \\
CogAgent-VQA & Vicuna-7B & \cellcolor{red!20} 1.67{\tiny (7)} & \cellcolor{blue!20} 1.97{\tiny (7)} & \cellcolor{red!25} 1.9{\tiny (6)} & \cellcolor{blue!40} 1.98{\tiny (3)} & \multicolumn{1}{>{\columncolor{red!10}}c|} {2.64{\tiny (9)}} & \cellcolor{blue!30} 2.03{\tiny (5)} & \cellcolor{red!20} 110.01{\tiny (7)} & \cellcolor{blue!20} 85.23{\tiny (7)} & \cellcolor{red!30} 86.75{\tiny (5)} & \cellcolor{blue!40} 84.33{\tiny (3)} & \multicolumn{1}{>{\columncolor{red!15}}c|} {53.99{\tiny (8)}} & \cellcolor{blue!15} 84.06{\tiny (8)} \\
MobileVLMV2-7B & Vicuna-7B & \cellcolor{red!25} 1.63{\tiny (6)} & \cellcolor{blue!15} 2.15{\tiny (8)} & \cellcolor{red!45} 1.62{\tiny (2)} & \cellcolor{blue!30} 2.11{\tiny (5)} & \multicolumn{1}{>{\columncolor{red!5}}c|} {2.92{\tiny (10)}} & \cellcolor{blue!15} 2.09{\tiny (8)} & \cellcolor{red!35} 114.26{\tiny (4)} & \cellcolor{blue!10} 75.68{\tiny (9)} & \cellcolor{red!40} 109.05{\tiny (3)} & \cellcolor{blue!30} 77.59{\tiny (5)} & \multicolumn{1}{>{\columncolor{red!5}}c|} {44.85{\tiny (10)}} & \cellcolor{blue!20} 84.29{\tiny (7)} \\
MoE-LLaVA-Phi2-2.7B & Phi2-2.7B & \cellcolor{red!45} 1.58{\tiny (2)} & \cellcolor{blue!50} 1.52{\tiny (1)} & \cellcolor{red!10} 2.1{\tiny (9)} & \cellcolor{blue!35} 2.0{\tiny (4)} & \multicolumn{1}{>{\columncolor{red!25}}c|} {2.49{\tiny (6)}} & \cellcolor{blue!40} 1.94{\tiny (3)} & \cellcolor{red!30} 114.0{\tiny (5)} & \cellcolor{blue!50} 120.55{\tiny (1)} & \cellcolor{red!10} 74.78{\tiny (9)} & \cellcolor{blue!35} 81.26{\tiny (4)} & \multicolumn{1}{>{\columncolor{red!20}}c|} {54.87{\tiny (7)}} & \cellcolor{blue!30} 89.09{\tiny (5)} \\
mPLUG-Owl2 & LLaMA2-7B & \cellcolor{red!15} 1.73{\tiny (8)} & \cellcolor{blue!35} 1.69{\tiny (4)} & \cellcolor{red!15} 2.07{\tiny (8)} & \cellcolor{blue!20} 2.43{\tiny (7)} & \multicolumn{1}{>{\columncolor{red!15}}c|} {2.62{\tiny (8)}} & \cellcolor{blue!10} 2.11{\tiny (9)} & \cellcolor{red!15} 103.62{\tiny (8)} & \cellcolor{blue!40} 106.26{\tiny (3)} & \cellcolor{red!15} 77.74{\tiny (8)} & \cellcolor{blue!15} 61.92{\tiny (8)} & \multicolumn{1}{>{\columncolor{red!10}}c|} {50.76{\tiny (9)}} & \cellcolor{blue!10} 80.06{\tiny (9)} \\
Qwen-VL-Chat & Qwen-7b & \cellcolor{red!5} 1.99{\tiny (10)} & \cellcolor{blue!5} 3.05{\tiny (10)} & \cellcolor{red!5} 2.45{\tiny (10)} & \cellcolor{blue!5} 2.95{\tiny (10)} & \multicolumn{1}{>{\columncolor{red!30}}c|} {2.39{\tiny (5)}} & \cellcolor{blue!5} 2.57{\tiny (10)} & \cellcolor{red!5} 87.91{\tiny (10)} & \cellcolor{blue!5} 43.56{\tiny (10)} & \cellcolor{red!5} 63.18{\tiny (10)} & \cellcolor{blue!5} 49.72{\tiny (10)} & \multicolumn{1}{>{\columncolor{red!30}}c|} {66.59{\tiny (5)}} & \cellcolor{blue!5} 62.19{\tiny (10)} \\ \hline

\end{tabular}

}
\vspace*{0.15cm}
\caption{Results of VLMs with LLM sizes ranging from 2.7B to 13B when LAC is adopted as the conformal score function. Each row contains a relative ranking among all models in the corresponding column (a small number in parentheses), and this ranking is also highlighted with color.}
\label{tab:lac}
\end{table*}

\begin{table*}[h]
\centering
\resizebox{\textwidth}{!}{%
\begin{tabular}{l|l|cccccc|cccccc}
\hline
\multirow{2}{*}{\textbf{VLM}} & \multirow{2}{*}{\textbf{LLM}} & \multicolumn{6}{c|}{\textbf{Coverage Rate (\%)}} & 
\multicolumn{6}{c}{\textbf{Acc (\%)} $\uparrow$} \\ 
\cline{3-14} 
&  & \textbf{MMB} & \textbf{OOD} & \textbf{SQA} & \textbf{SB} & \multicolumn{1}{c|}{\textbf{AI2D}} & \textbf{Avg.} & \textbf{MMB}  & \textbf{OOD} & \textbf{SQA} & \textbf{SB} & \multicolumn{1}{c|}{\textbf{AI2D}} & \textbf{Avg.} \\ \hline 
LLaVA-v1.6-13B & Vicuna-13b & 98.99 & 96.20 & 99.29 & 97.36 & 98.86 & 98.14 & \cellcolor{red!45} 76.75{\tiny (2)} & \cellcolor{blue!30} 72.93{\tiny (5)} & \cellcolor{red!35} 70.56{\tiny (4)} & \cellcolor{blue!50} 70.37{\tiny (1)} & \multicolumn{1}{>{\columncolor{red!50}}c|} {73.67{\tiny (1)}} & \cellcolor{blue!50} 72.85{\tiny (1)} \\
Monkey-Chat & Qwen-7b & 97.85 & 96.27 & 98.84 & 96.50 & 97.28 & 97.35 & \cellcolor{red!50} 76.98{\tiny (1)} & \cellcolor{blue!25} 70.6{\tiny (6)} & \cellcolor{red!45} 74.66{\tiny (2)} & \cellcolor{blue!20} 66.1{\tiny (7)} & \multicolumn{1}{>{\columncolor{red!40}}c|} {67.95{\tiny (3)}} & \cellcolor{blue!45} 71.26{\tiny (2)} \\
LLaVA-v1.6-7B & Vicuna-7B & 98.45 & 97.89 & 97.88 & 96.74 & 96.19 & 97.43 & \cellcolor{red!35} 75.56{\tiny (4)} & \cellcolor{blue!40} 73.7{\tiny (3)} & \cellcolor{red!20} 65.86{\tiny (7)} & \cellcolor{blue!45} 69.06{\tiny (2)} & \multicolumn{1}{>{\columncolor{red!45}}c|} {69.75{\tiny (2)}} & \cellcolor{blue!40} 70.78{\tiny (3)} \\
InternLM-XComposer2-VL & InternLM-7b & 96.57 & 92.48 & 98.74 & 94.46 & 96.28 & 95.71 & \cellcolor{red!10} 71.77{\tiny (9)} & \cellcolor{blue!20} 70.04{\tiny (7)} & \cellcolor{red!50} 77.95{\tiny (1)} & \cellcolor{blue!15} 64.44{\tiny (8)} & \multicolumn{1}{>{\columncolor{red!35}}c|} {66.13{\tiny (4)}} & \cellcolor{blue!35} 70.07{\tiny (4)} \\
Yi-VL-6B & Yi-6B & 98.63 & 95.43 & 98.13 & 95.94 & 96.77 & 96.98 & \cellcolor{red!30} 75.24{\tiny (5)} & \cellcolor{blue!45} 73.91{\tiny (2)} & \cellcolor{red!25} 66.72{\tiny (6)} & \cellcolor{blue!25} 66.25{\tiny (6)} & \multicolumn{1}{>{\columncolor{red!25}}c|} {58.84{\tiny (6)}} & \cellcolor{blue!30} 68.19{\tiny (5)} \\
CogAgent-VQA & Vicuna-7B & 98.54 & 95.64 & 97.47 & 95.94 & 93.83 & 96.28 & \cellcolor{red!25} 74.78{\tiny (6)} & \cellcolor{blue!15} 68.57{\tiny (8)} & \cellcolor{red!30} 67.12{\tiny (5)} & \cellcolor{blue!40} 68.01{\tiny (3)} & \multicolumn{1}{>{\columncolor{red!20}}c|} {58.2{\tiny (7)}} & \cellcolor{blue!25} 67.34{\tiny (6)} \\
MobileVLMV2-7B & Vicuna-7B & 97.99 & 96.27 & 99.04 & 97.67 & 95.87 & 97.37 & \cellcolor{red!40} 75.97{\tiny (3)} & \cellcolor{blue!10} 66.53{\tiny (9)} & \cellcolor{red!40} 72.33{\tiny (3)} & \cellcolor{blue!35} 66.71{\tiny (4)} & \multicolumn{1}{>{\columncolor{red!5}}c|} {53.55{\tiny (10)}} & \cellcolor{blue!20} 67.02{\tiny (7)} \\
MoE-LLaVA-Phi2-2.7B & Phi2-2.7B & 99.50 & 93.95 & 97.07 & 97.60 & 96.50 & 96.92 & \cellcolor{red!20} 73.73{\tiny (7)} & \cellcolor{blue!50} 74.82{\tiny (1)} & \cellcolor{red!10} 64.04{\tiny (9)} & \cellcolor{blue!30} 66.42{\tiny (5)} & \multicolumn{1}{>{\columncolor{red!15}}c|} {55.76{\tiny (8)}} & \cellcolor{blue!15} 66.95{\tiny (8)} \\
mPLUG-Owl2 & LLaMA2-7B & 99.27 & 95.08 & 98.18 & 97.09 & 95.81 & 97.08 & \cellcolor{red!15} 73.05{\tiny (8)} & \cellcolor{blue!35} 73.28{\tiny (4)} & \cellcolor{red!15} 65.71{\tiny (8)} & \cellcolor{blue!10} 61.49{\tiny (9)} & \multicolumn{1}{>{\columncolor{red!10}}c|} {54.38{\tiny (9)}} & \cellcolor{blue!10} 65.58{\tiny (9)} \\
Qwen-VL-Chat & Qwen-7b & 96.21 & 93.46 & 92.01 & 92.97 & 96.70 & 94.27 & \cellcolor{red!5} 71.4{\tiny (10)} & \cellcolor{blue!5} 54.22{\tiny (10)} & \cellcolor{red!5} 63.23{\tiny (10)} & \cellcolor{blue!5} 59.79{\tiny (10)} & \multicolumn{1}{>{\columncolor{red!30}}c|} {65.09{\tiny (5)}} & \cellcolor{blue!5} 62.74{\tiny (10)} \\ \hline

\multirow{2}{*}{\textbf{VLM}} & \multirow{2}{*}{\textbf{LLM}} & \multicolumn{6}{c|}{\textbf{SS} $\downarrow$} & 
\multicolumn{6}{c}{\textbf{UAcc (\%)} $\uparrow$}  \\ 
\cline{3-14} 
&  & \textbf{MMB} & \textbf{OOD} & \textbf{SQA} & \textbf{SB} & \multicolumn{1}{c|}{\textbf{AI2D}} & \textbf{Avg.} & \textbf{MMB}  & \textbf{OOD} & \textbf{SQA} & \textbf{SB} & \multicolumn{1}{c|}{\textbf{AI2D}} & \textbf{Avg.} \\ \hline 
LLaVA-v1.6-13B & Vicuna-13b & \cellcolor{red!45} 3.13{\tiny (2)} & \cellcolor{blue!30} 2.69{\tiny (5)} & \cellcolor{red!40} 3.22{\tiny (3)} & \cellcolor{blue!35} 3.13{\tiny (4)} & \multicolumn{1}{>{\columncolor{red!45}}c|} {3.02{\tiny (2)}} & \cellcolor{blue!45} 3.04{\tiny (2)} & \cellcolor{red!45} 60.11{\tiny (2)} & \cellcolor{blue!30} 66.51{\tiny (5)} & \cellcolor{red!40} 53.71{\tiny (3)} & \cellcolor{blue!45} 55.1{\tiny (2)} & \multicolumn{1}{>{\columncolor{red!50}}c|} {59.82{\tiny (1)}} & \cellcolor{blue!50} 59.05{\tiny (1)} \\
Monkey-Chat & Qwen-7b & \cellcolor{red!5} 3.79{\tiny (10)} & \cellcolor{blue!5} 3.67{\tiny (10)} & \cellcolor{red!15} 3.45{\tiny (8)} & \cellcolor{blue!5} 4.01{\tiny (10)} & \multicolumn{1}{>{\columncolor{red!5}}c|} {4.04{\tiny (10)}} & \cellcolor{blue!5} 3.79{\tiny (10)} & \cellcolor{red!5} 49.79{\tiny (10)} & \cellcolor{blue!10} 47.14{\tiny (9)} & \cellcolor{red!30} 52.93{\tiny (5)} & \cellcolor{blue!10} 40.35{\tiny (9)} & \multicolumn{1}{>{\columncolor{red!20}}c|} {41.2{\tiny (7)}} & \cellcolor{blue!10} 46.28{\tiny (9)} \\
LLaVA-v1.6-7B & Vicuna-7B & \cellcolor{red!40} 3.15{\tiny (3)} & \cellcolor{blue!20} 2.96{\tiny (7)} & \cellcolor{red!45} 3.03{\tiny (2)} & \cellcolor{blue!45} 3.11{\tiny (2)} & \multicolumn{1}{>{\columncolor{red!50}}c|} {2.98{\tiny (1)}} & \cellcolor{blue!40} 3.05{\tiny (3)} & \cellcolor{red!40} 58.68{\tiny (3)} & \cellcolor{blue!25} 60.96{\tiny (6)} & \cellcolor{red!35} 53.23{\tiny (4)} & \cellcolor{blue!40} 54.39{\tiny (3)} & \multicolumn{1}{>{\columncolor{red!45}}c|} {57.42{\tiny (2)}} & \cellcolor{blue!45} 56.94{\tiny (2)} \\
InternLM-XComposer2-VL & InternLM-7b & \cellcolor{red!15} 3.48{\tiny (8)} & \cellcolor{blue!35} 2.57{\tiny (4)} & \cellcolor{red!20} 3.38{\tiny (7)} & \cellcolor{blue!15} 3.58{\tiny (8)} & \multicolumn{1}{>{\columncolor{red!20}}c|} {3.67{\tiny (7)}} & \cellcolor{blue!20} 3.34{\tiny (7)} & \cellcolor{red!10} 50.53{\tiny (9)} & \cellcolor{blue!35} 66.64{\tiny (4)} & \cellcolor{red!45} 56.43{\tiny (2)} & \cellcolor{blue!20} 44.11{\tiny (7)} & \multicolumn{1}{>{\columncolor{red!40}}c|} {44.11{\tiny (3)}} & \cellcolor{blue!25} 52.36{\tiny (6)} \\
Yi-VL-6B & Yi-6B & \cellcolor{red!35} 3.33{\tiny (4)} & \cellcolor{blue!40} 2.51{\tiny (3)} & \cellcolor{red!10} 3.5{\tiny (9)} & \cellcolor{blue!40} 3.12{\tiny (3)} & \multicolumn{1}{>{\columncolor{red!25}}c|} {3.49{\tiny (6)}} & \cellcolor{blue!30} 3.19{\tiny (5)} & \cellcolor{red!35} 55.42{\tiny (4)} & \cellcolor{blue!45} 72.23{\tiny (2)} & \cellcolor{red!10} 46.66{\tiny (9)} & \cellcolor{blue!35} 52.08{\tiny (4)} & \multicolumn{1}{>{\columncolor{red!25}}c|} {41.28{\tiny (6)}} & \cellcolor{blue!30} 53.53{\tiny (5)} \\
CogAgent-VQA & Vicuna-7B & \cellcolor{red!50} 3.0{\tiny (1)} & \cellcolor{blue!25} 2.94{\tiny (6)} & \cellcolor{red!50} 2.83{\tiny (1)} & \cellcolor{blue!50} 3.0{\tiny (1)} & \multicolumn{1}{>{\columncolor{red!40}}c|} {3.24{\tiny (3)}} & \cellcolor{blue!50} 3.0{\tiny (1)} & \cellcolor{red!50} 61.12{\tiny (1)} & \cellcolor{blue!20} 57.05{\tiny (7)} & \cellcolor{red!50} 58.04{\tiny (1)} & \cellcolor{blue!50} 55.6{\tiny (1)} & \multicolumn{1}{>{\columncolor{red!35}}c|} {44.01{\tiny (4)}} & \cellcolor{blue!40} 55.16{\tiny (3)} \\
MobileVLMV2-7B & Vicuna-7B & \cellcolor{red!20} 3.44{\tiny (7)} & \cellcolor{blue!15} 3.07{\tiny (8)} & \cellcolor{red!5} 3.61{\tiny (10)} & \cellcolor{blue!20} 3.49{\tiny (7)} & \multicolumn{1}{>{\columncolor{red!10}}c|} {3.87{\tiny (9)}} & \cellcolor{blue!15} 3.5{\tiny (8)} & \cellcolor{red!30} 54.12{\tiny (5)} & \cellcolor{blue!15} 53.01{\tiny (8)} & \cellcolor{red!25} 49.09{\tiny (6)} & \cellcolor{blue!25} 46.77{\tiny (6)} & \multicolumn{1}{>{\columncolor{red!5}}c|} {33.93{\tiny (10)}} & \cellcolor{blue!15} 47.38{\tiny (8)} \\
MoE-LLaVA-Phi2-2.7B & Phi2-2.7B & \cellcolor{red!10} 3.5{\tiny (9)} & \cellcolor{blue!50} 2.27{\tiny (1)} & \cellcolor{red!35} 3.3{\tiny (4)} & \cellcolor{blue!30} 3.38{\tiny (5)} & \multicolumn{1}{>{\columncolor{red!35}}c|} {3.34{\tiny (4)}} & \cellcolor{blue!35} 3.16{\tiny (4)} & \cellcolor{red!20} 51.66{\tiny (7)} & \cellcolor{blue!50} 80.91{\tiny (1)} & \cellcolor{red!15} 47.58{\tiny (8)} & \cellcolor{blue!30} 48.08{\tiny (5)} & \multicolumn{1}{>{\columncolor{red!15}}c|} {40.86{\tiny (8)}} & \cellcolor{blue!35} 53.82{\tiny (4)} \\
mPLUG-Owl2 & LLaMA2-7B & \cellcolor{red!30} 3.36{\tiny (5)} & \cellcolor{blue!45} 2.49{\tiny (2)} & \cellcolor{red!30} 3.35{\tiny (5)} & \cellcolor{blue!25} 3.43{\tiny (6)} & \multicolumn{1}{>{\columncolor{red!30}}c|} {3.38{\tiny (5)}} & \cellcolor{blue!25} 3.2{\tiny (6)} & \cellcolor{red!25} 53.18{\tiny (6)} & \cellcolor{blue!40} 72.22{\tiny (3)} & \cellcolor{red!20} 48.1{\tiny (7)} & \cellcolor{blue!15} 43.89{\tiny (8)} & \multicolumn{1}{>{\columncolor{red!10}}c|} {39.42{\tiny (9)}} & \cellcolor{blue!20} 51.36{\tiny (7)} \\
Qwen-VL-Chat & Qwen-7b & \cellcolor{red!25} 3.41{\tiny (6)} & \cellcolor{blue!10} 3.59{\tiny (9)} & \cellcolor{red!30} 3.35{\tiny (5)} & \cellcolor{blue!10} 3.69{\tiny (9)} & \multicolumn{1}{>{\columncolor{red!15}}c|} {3.8{\tiny (8)}} & \cellcolor{blue!10} 3.57{\tiny (9)} & \cellcolor{red!15} 51.25{\tiny (8)} & \cellcolor{blue!5} 37.01{\tiny (10)} & \cellcolor{red!5} 46.24{\tiny (10)} & \cellcolor{blue!5} 39.67{\tiny (10)} & \multicolumn{1}{>{\columncolor{red!30}}c|} {42.0{\tiny (5)}} & \cellcolor{blue!5} 43.23{\tiny (10)} \\ \hline

\end{tabular}

}
\vspace*{0.15cm}
\caption{Results of VLMs with LLM sizes ranging from 2.7B to 13B when APS is adopted as the conformal score function. Each row contains relative ranking among all models in corresponding column (small number in parentheses) and this ranking also highlited with color.}
\label{tab:aps}
\end{table*}

\section{Detailed Results of LAC and APS}
Here, we present individual results for each of the LAC and APS score functions. These results vary for both approaches.

Table ~\ref{tab:lac} contains the results for the LAC score function. The coverage rate frequently falls below 90\%, indicating that the coverage guarantee requirement sometimes has not been met. However, the average values of coverage rate either exceed 90\% or are below but very close to this value. In contrast, the coverage rates of the APS method presented in Table ~\ref{tab:aps} for each model considered significantly exceed the 90\% threshold on each dataset. This difference highlights how crucial it is to combine both methods and average them for a thorough evaluation. 

The obtained Set Sizes values also vary significantly  Overall, APS score typically generates larger prediction sets. For LAC, the averaged values do not exceed 2.51 among all datasets, while the smallest average value for the APS method was 2.87 and exceeded 3 for all other models.

Moreover, APS can also result in a notably distinct ranking of VLMs in terms of uncertainty compared to LAC.
Noticeably, the Monkey-Chat ranked eighth when using the LAC score function but has the lowest Set Sizes when using APS. 
Similarly, the CogVLM-Chat ranks ninth when LAC is utilized for uncertainty quantification while achieving a significantly higher fourth rank when utilizing APS.

\section{Effects of LLM Scale} \label{app:scale}

\begin{figure*}[h]
  \centering
  \includegraphics[width=\linewidth]{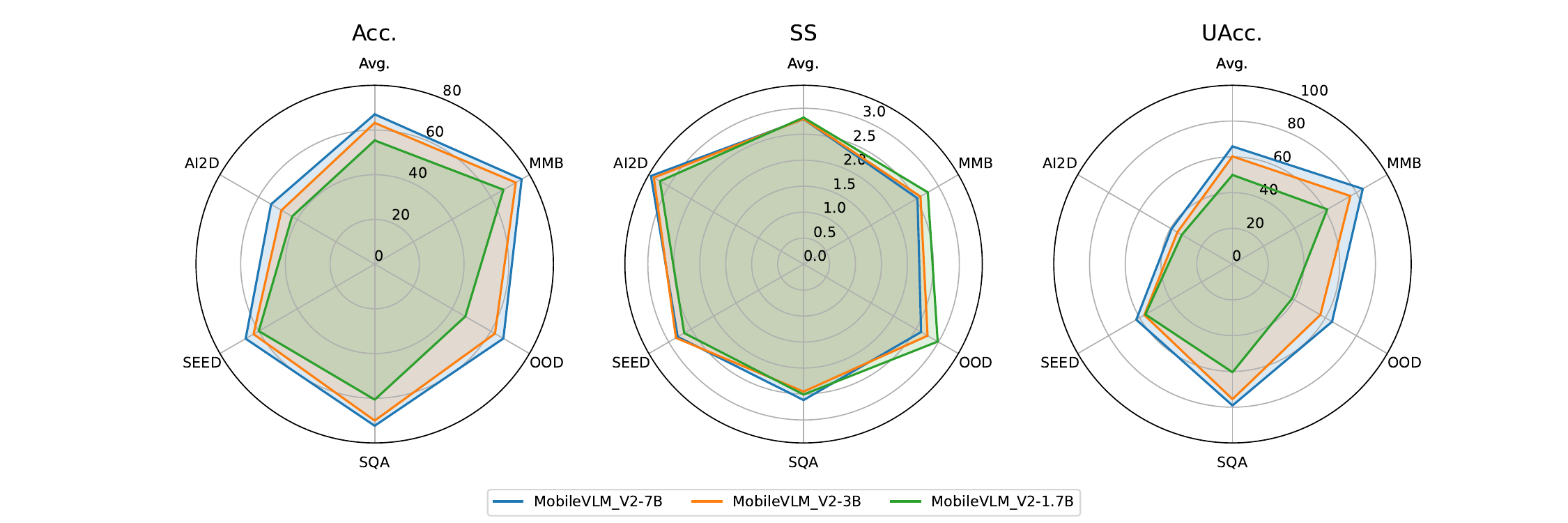}
  \caption{Comparison of MobileVLMV2 with LLM of different size.}
  \label{fig:scale_movilevlm2_2}
\end{figure*}

\begin{figure*}[h]
  \centering
  \includegraphics[width=\linewidth]{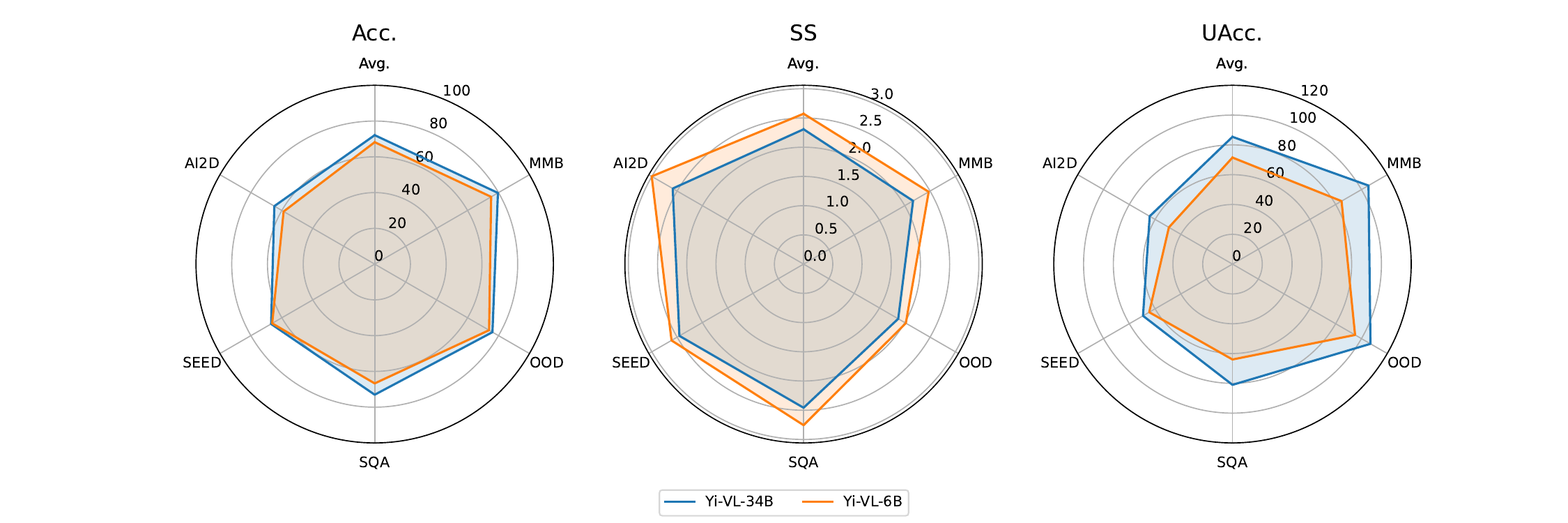}
  \caption{Comparison of Yi-VL with LLM of different size.}
  \label{fig:scale_yi_vl}
\end{figure*}

Both MobileVLMV2 and Yi-VL model series comparisons are presented in Figures ~\ref{fig:scale_movilevlm2_2} and ~\ref{fig:scale_yi_vl}, respectively. 
Accuracy increases as the corresponding LLM size grows for both model series.
We also observe Set Sizes decrease when moving from Yi-VL-6B to Yi-VL-34B. 

However, we do not observe a similar situation when LLM in MobileVLMV2 grows. The average values across the five tasks are very close to each other. All three models alternately show the highest Set Sizes: MobileVLMV2-1.7B demonstrates the largest Set Sizes in MMBench and OODCV, MobileVLMV2-3B on SEEDBench, MobileVLMV2-7B - on ScienceQA and AI2D. Therefore, we can conclude that, at least for models that are not very large, the uncertainty of VLMs does not necessarily decrease with increasing LLM size.

\section{Effects of Chat Finetuning} 
\label{sec:base_vs_chat}
\begin{figure*}[h]
  \centering
  \includegraphics[width=\linewidth]{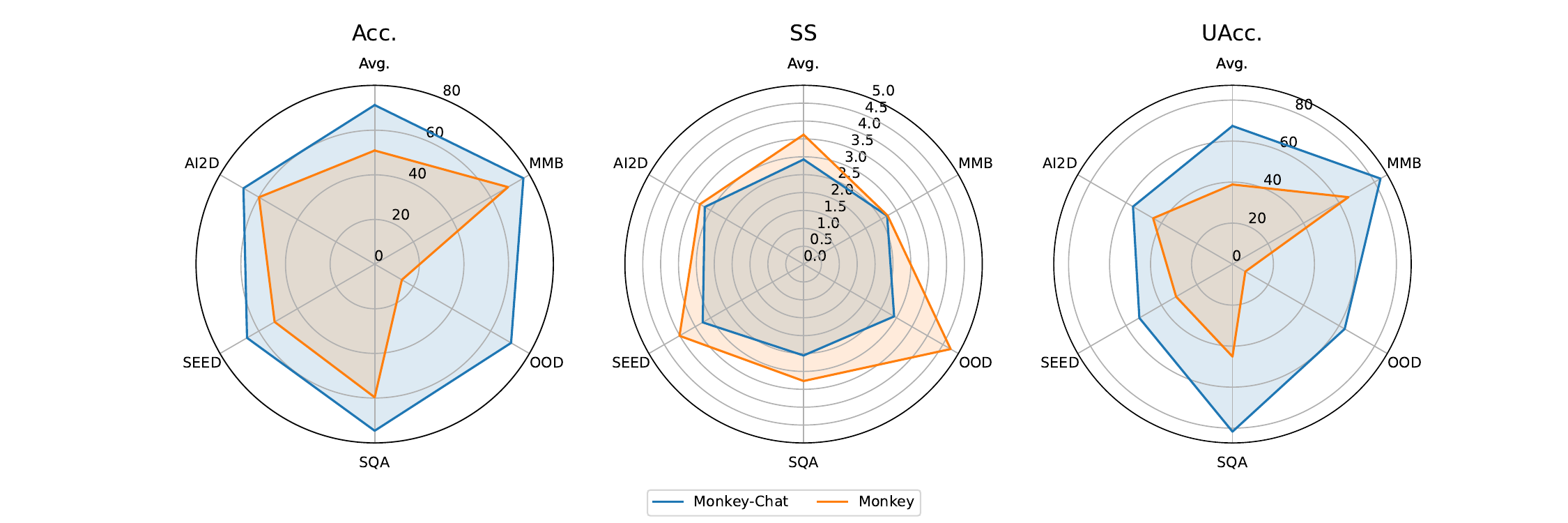}
  \caption{Comparison of Monkey and Monkey-Chat.}
  \label{fig:base_vs_chat_monkey}
\end{figure*}

\begin{figure*}[h]
  \centering
  \includegraphics[width=\linewidth]{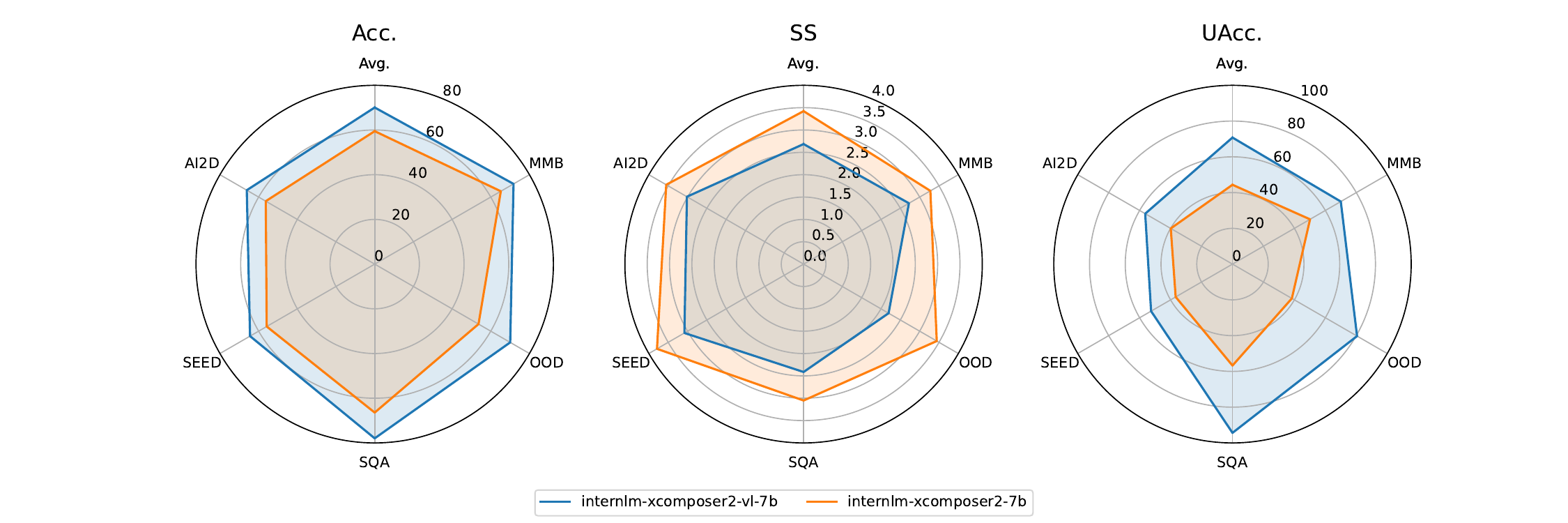}
  \caption{Comparison of InternLM-XComposer2 and InternLM-XComposer2-VL(Chat).}
  \label{fig:base_vs_chat_intern}
\end{figure*}

In Figures ~\ref{fig:base_vs_chat_monkey}-~\ref{fig:base_vs_chat_intern}
we present additional comparisons between the base and chat-tuned models. Unlike the Qwen model, we observed significantly lower uncertainty for chat versions within model series Monkey and  InternLM-XComposer2 when comparing them with base versions in terms of Set Sizes. 

In most cases, Chat Fine-Tuning leads to an uncertainty decrease for VLMs since base VLMs have not seen enough visual question-answer pairs.

\section{SEEDBench Detailed Results}
\label{app:seed}

Originally SEEDBench has 9 dimensions covers image modality. In Table ~\ref{tab:seedbench} we provide five dimensions with the most questions:
\begin{itemize}
    \item Scene Understanding
    \item Instance Identity
    \item Instance Location
    \item Instance Attribute
    \item Instances Counting
\end{itemize}
Scene Understanding was the easiest task in terms of accuracy (each model surpassed 70\%), and the average uncertainty across all models was the lowest among the five tasks. On the contrary Instance Attribute and Instances Counting are the most difficult tasks; the best model in both tasks (LLaVA-v1.6-13B) does not exceed 64\% in accuracy. The highest uncertainty is also observed in these tasks.

\begin{table*}[t]
\centering
\resizebox{\textwidth}{!}{%
\begin{tabular}{l|l|cccccc|cccccc}
\hline
\multirow{2}{*}{\textbf{VLM}} & \multirow{2}{*}{\textbf{LLM}} & \multicolumn{6}{c|}{\textbf{Coverage Rate (\%)}} & 
\multicolumn{6}{c}{\textbf{Acc (\%)} $\uparrow$} \\ 
\cline{3-14} 
&  & \textbf{SU} & \textbf{II} & \textbf{IL} & \textbf{IA} & \multicolumn{1}{c|}{\textbf{IC}} & \textbf{Avg.} & \textbf{SU}  & \textbf{II} & \textbf{IL} & \textbf{IA} & \multicolumn{1}{c|}{\textbf{IC}} & \textbf{Avg.} \\ \hline 
LLaVA-v1.6-13B & Vicuna-13b & 93.73 & 92.74 & 93.35 & 91.31 & 92.32 & 92.69 & \cellcolor{red!50} 77.07{\tiny (1)} & \cellcolor{blue!45} 75.0{\tiny (2)} & \cellcolor{red!50} 74.62{\tiny (1)} & \cellcolor{blue!50} 63.39{\tiny (1)} & \multicolumn{1}{>{\columncolor{red!50}}c|} {63.48{\tiny (1)}} & \cellcolor{blue!50} 70.71{\tiny (1)} \\
Monkey-Chat & Qwen-7b & 93.29 & 93.01 & 92.80 & 93.05 & 94.08 & 93.24 & \cellcolor{red!20} 73.21{\tiny (7)} & \cellcolor{blue!20} 65.72{\tiny (7)} & \cellcolor{red!35} 70.49{\tiny (4)} & \cellcolor{blue!30} 60.12{\tiny (5)} & \multicolumn{1}{>{\columncolor{red!35}}c|} {59.97{\tiny (4)}} & \cellcolor{blue!25} 65.9{\tiny (6)} \\
LLaVA-v1.6-7B & Vicuna-7B & 94.02 & 93.45 & 94.00 & 90.49 & 92.57 & 92.90 & \cellcolor{red!45} 74.79{\tiny (2)} & \cellcolor{blue!40} 72.16{\tiny (3)} & \cellcolor{red!45} 73.63{\tiny (2)} & \cellcolor{blue!30} 60.12{\tiny (5)} & \multicolumn{1}{>{\columncolor{red!40}}c|} {61.36{\tiny (3)}} & \cellcolor{blue!45} 68.41{\tiny (2)} \\
InternLM-XComposer2-VL & InternLM-7b & 92.84 & 93.40 & 92.00 & 92.54 & 91.58 & 92.47 & \cellcolor{red!35} 74.29{\tiny (4)} & \cellcolor{blue!10} 64.41{\tiny (9)} & \cellcolor{red!20} 66.67{\tiny (7)} & \cellcolor{blue!20} 58.28{\tiny (7)} & \multicolumn{1}{>{\columncolor{red!20}}c|} {56.7{\tiny (7)}} & \cellcolor{blue!15} 64.07{\tiny (8)} \\
Yi-VL-6B & Yi-6B & 93.83 & 92.69 & 93.23 & 91.41 & 91.63 & 92.55 & \cellcolor{red!15} 72.51{\tiny (8)} & \cellcolor{blue!25} 68.34{\tiny (6)} & \cellcolor{red!25} 68.17{\tiny (6)} & \cellcolor{blue!40} 61.55{\tiny (3)} & \multicolumn{1}{>{\columncolor{red!30}}c|} {59.4{\tiny (5)}} & \cellcolor{blue!30} 66.0{\tiny (5)} \\
CogAgent-VQA & Vicuna-7B & 92.46 & 93.34 & 93.53 & 89.67 & 91.46 & 92.09 & \cellcolor{red!45} 74.79{\tiny (2)} & \cellcolor{blue!50} 75.55{\tiny (1)} & \cellcolor{red!40} 72.69{\tiny (3)} & \cellcolor{blue!15} 55.83{\tiny (8)} & \multicolumn{1}{>{\columncolor{red!25}}c|} {58.33{\tiny (6)}} & \cellcolor{blue!40} 67.44{\tiny (3)} \\
MobileVLMV2-7B & Vicuna-7B & 93.79 & 94.00 & 93.59 & 94.27 & 93.75 & 93.88 & \cellcolor{red!25} 73.53{\tiny (6)} & \cellcolor{blue!35} 70.96{\tiny (4)} & \cellcolor{red!15} 66.28{\tiny (8)} & \cellcolor{blue!45} 61.96{\tiny (2)} & \multicolumn{1}{>{\columncolor{red!45}}c|} {63.15{\tiny (2)}} & \cellcolor{blue!35} 67.18{\tiny (4)} \\
MoE-LLaVA-Phi2-2.7B & Phi2-2.7B & 94.74 & 93.12 & 93.16 & 88.55 & 92.24 & 92.36 & \cellcolor{red!30} 74.03{\tiny (5)} & \cellcolor{blue!30} 69.76{\tiny (5)} & \cellcolor{red!30} 68.47{\tiny (5)} & \cellcolor{blue!30} 60.12{\tiny (5)} & \multicolumn{1}{>{\columncolor{red!10}}c|} {55.56{\tiny (9)}} & \cellcolor{blue!20} 65.59{\tiny (7)} \\
mPLUG-Owl2 & LLaMA2-7B & 93.57 & 93.72 & 93.51 & 91.00 & 91.63 & 92.69 & \cellcolor{red!10} 71.82{\tiny (9)} & \cellcolor{blue!15} 64.63{\tiny (8)} & \cellcolor{red!5} 61.72{\tiny (10)} & \cellcolor{blue!10} 52.97{\tiny (9)} & \multicolumn{1}{>{\columncolor{red!15}}c|} {55.88{\tiny (8)}} & \cellcolor{blue!10} 61.4{\tiny (9)} \\
Qwen-VL-Chat & Qwen-7b & 91.64 & 89.63 & 91.46 & 88.45 & 91.09 & 90.45 & \cellcolor{red!5} 71.37{\tiny (10)} & \cellcolor{blue!5} 61.68{\tiny (10)} & \cellcolor{red!10} 62.97{\tiny (9)} & \cellcolor{blue!5} 51.53{\tiny (10)} & \multicolumn{1}{>{\columncolor{red!5}}c|} {51.72{\tiny (10)}} & \cellcolor{blue!5} 59.85{\tiny (10)} \\ \hline

\multirow{2}{*}{\textbf{VLM}} & \multirow{2}{*}{\textbf{LLM}} & \multicolumn{6}{c|}{\textbf{Coverage Rate (\%)}} & 
\multicolumn{6}{c}{\textbf{UAcc (\%)} $\uparrow$} \\ 
\cline{3-14} 
&  & \textbf{SU} & \textbf{II} & \textbf{IL} & \textbf{IA} & \multicolumn{1}{c|}{\textbf{IC}} & \textbf{Avg.} & \textbf{SU}  & \textbf{II} & \textbf{IL} & \textbf{IA} & \multicolumn{1}{c|}{\textbf{IC}} & \textbf{Avg.} \\ \hline  
LLaVA-v1.6-13B & Vicuna-13b & \cellcolor{red!50} 2.21{\tiny (1)} & \cellcolor{blue!45} 2.37{\tiny (2)} & \cellcolor{red!50} 2.29{\tiny (1)} & \cellcolor{blue!40} 2.67{\tiny (3)} & \multicolumn{1}{>{\columncolor{red!40}}c|} {2.67{\tiny (3)}} & \cellcolor{blue!50} 2.44{\tiny (1)} & \cellcolor{red!50} 95.26{\tiny (1)} & \cellcolor{blue!50} 84.8{\tiny (1)} & \cellcolor{red!50} 87.01{\tiny (1)} & \cellcolor{blue!50} 59.75{\tiny (1)} & \multicolumn{1}{>{\columncolor{red!45}}c|} {59.14{\tiny (2)}} & \cellcolor{blue!50} 77.19{\tiny (1)} \\
Monkey-Chat & Qwen-7b & \cellcolor{red!5} 2.88{\tiny (10)} & \cellcolor{blue!5} 3.32{\tiny (10)} & \cellcolor{red!10} 3.13{\tiny (9)} & \cellcolor{blue!10} 3.51{\tiny (9)} & \multicolumn{1}{>{\columncolor{red!10}}c|} {3.43{\tiny (9)}} & \cellcolor{blue!10} 3.25{\tiny (9)} & \cellcolor{red!10} 68.42{\tiny (9)} & \cellcolor{blue!10} 50.91{\tiny (9)} & \cellcolor{red!20} 57.49{\tiny (7)} & \cellcolor{blue!15} 42.72{\tiny (8)} & \multicolumn{1}{>{\columncolor{red!15}}c|} {43.45{\tiny (8)}} & \cellcolor{blue!10} 52.6{\tiny (9)} \\
LLaVA-v1.6-7B & Vicuna-7B & \cellcolor{red!40} 2.31{\tiny (3)} & \cellcolor{blue!40} 2.48{\tiny (3)} & \cellcolor{red!40} 2.38{\tiny (3)} & \cellcolor{blue!45} 2.63{\tiny (2)} & \multicolumn{1}{>{\columncolor{red!50}}c|} {2.61{\tiny (1)}} & \cellcolor{blue!45} 2.48{\tiny (2)} & \cellcolor{red!35} 85.44{\tiny (4)} & \cellcolor{blue!40} 76.21{\tiny (3)} & \cellcolor{red!45} 81.93{\tiny (2)} & \cellcolor{blue!35} 56.64{\tiny (4)} & \multicolumn{1}{>{\columncolor{red!40}}c|} {58.3{\tiny (3)}} & \cellcolor{blue!45} 71.7{\tiny (2)} \\
InternLM-XComposer2-VL & InternLM-7b & \cellcolor{red!15} 2.58{\tiny (8)} & \cellcolor{blue!10} 3.15{\tiny (9)} & \cellcolor{red!15} 3.02{\tiny (8)} & \cellcolor{blue!10} 3.51{\tiny (9)} & \multicolumn{1}{>{\columncolor{red!15}}c|} {3.38{\tiny (8)}} & \cellcolor{blue!15} 3.13{\tiny (8)} & \cellcolor{red!15} 76.55{\tiny (8)} & \cellcolor{blue!15} 50.93{\tiny (8)} & \cellcolor{red!15} 55.19{\tiny (8)} & \cellcolor{blue!10} 41.03{\tiny (9)} & \multicolumn{1}{>{\columncolor{red!10}}c|} {41.21{\tiny (9)}} & \cellcolor{blue!15} 52.98{\tiny (8)} \\
Yi-VL-6B & Yi-6B & \cellcolor{red!20} 2.49{\tiny (7)} & \cellcolor{blue!30} 2.62{\tiny (5)} & \cellcolor{red!30} 2.55{\tiny (5)} & \cellcolor{blue!30} 2.7{\tiny (5)} & \multicolumn{1}{>{\columncolor{red!45}}c|} {2.62{\tiny (2)}} & \cellcolor{blue!35} 2.6{\tiny (4)} & \cellcolor{red!25} 78.63{\tiny (6)} & \cellcolor{blue!30} 66.87{\tiny (5)} & \cellcolor{red!30} 68.03{\tiny (5)} & \cellcolor{blue!40} 57.07{\tiny (3)} & \multicolumn{1}{>{\columncolor{red!35}}c|} {55.81{\tiny (4)}} & \cellcolor{blue!30} 65.28{\tiny (5)} \\
CogAgent-VQA & Vicuna-7B & \cellcolor{red!50} 2.21{\tiny (1)} & \cellcolor{blue!50} 2.34{\tiny (1)} & \cellcolor{red!45} 2.33{\tiny (2)} & \cellcolor{blue!35} 2.68{\tiny (4)} & \multicolumn{1}{>{\columncolor{red!25}}c|} {2.87{\tiny (6)}} & \cellcolor{blue!40} 2.49{\tiny (3)} & \cellcolor{red!45} 89.91{\tiny (2)} & \cellcolor{blue!45} 84.69{\tiny (2)} & \cellcolor{red!40} 81.7{\tiny (3)} & \cellcolor{blue!25} 51.6{\tiny (6)} & \multicolumn{1}{>{\columncolor{red!25}}c|} {49.94{\tiny (6)}} & \cellcolor{blue!40} 71.57{\tiny (3)} \\
MobileVLMV2-7B & Vicuna-7B & \cellcolor{red!35} 2.42{\tiny (4)} & \cellcolor{blue!25} 2.82{\tiny (6)} & \cellcolor{red!25} 2.92{\tiny (6)} & \cellcolor{blue!25} 2.95{\tiny (6)} & \multicolumn{1}{>{\columncolor{red!40}}c|} {2.67{\tiny (3)}} & \cellcolor{blue!25} 2.76{\tiny (6)} & \cellcolor{red!30} 83.69{\tiny (5)} & \cellcolor{blue!25} 66.01{\tiny (6)} & \cellcolor{red!25} 58.62{\tiny (6)} & \cellcolor{blue!30} 53.35{\tiny (5)} & \multicolumn{1}{>{\columncolor{red!50}}c|} {59.89{\tiny (1)}} & \cellcolor{blue!25} 64.31{\tiny (6)} \\
MoE-LLaVA-Phi2-2.7B & Phi2-2.7B & \cellcolor{red!25} 2.46{\tiny (6)} & \cellcolor{blue!35} 2.57{\tiny (4)} & \cellcolor{red!35} 2.53{\tiny (4)} & \cellcolor{blue!50} 2.55{\tiny (1)} & \multicolumn{1}{>{\columncolor{red!20}}c|} {2.88{\tiny (7)}} & \cellcolor{blue!35} 2.6{\tiny (4)} & \cellcolor{red!40} 85.97{\tiny (3)} & \cellcolor{blue!35} 73.71{\tiny (4)} & \cellcolor{red!35} 70.87{\tiny (4)} & \cellcolor{blue!45} 58.99{\tiny (2)} & \multicolumn{1}{>{\columncolor{red!20}}c|} {47.34{\tiny (7)}} & \cellcolor{blue!35} 67.38{\tiny (4)} \\
mPLUG-Owl2 & LLaMA2-7B & \cellcolor{red!30} 2.44{\tiny (5)} & \cellcolor{blue!20} 3.03{\tiny (7)} & \cellcolor{red!20} 2.98{\tiny (7)} & \cellcolor{blue!20} 2.96{\tiny (7)} & \multicolumn{1}{>{\columncolor{red!30}}c|} {2.73{\tiny (5)}} & \cellcolor{blue!20} 2.83{\tiny (7)} & \cellcolor{red!20} 76.71{\tiny (7)} & \cellcolor{blue!20} 53.33{\tiny (7)} & \cellcolor{red!10} 51.86{\tiny (9)} & \cellcolor{blue!20} 44.46{\tiny (7)} & \multicolumn{1}{>{\columncolor{red!30}}c|} {50.27{\tiny (5)}} & \cellcolor{blue!20} 55.33{\tiny (7)} \\
Qwen-VL-Chat & Qwen-7b & \cellcolor{red!10} 2.82{\tiny (9)} & \cellcolor{blue!15} 3.08{\tiny (8)} & \cellcolor{red!5} 3.3{\tiny (10)} & \cellcolor{blue!15} 3.3{\tiny (8)} & \multicolumn{1}{>{\columncolor{red!5}}c|} {3.93{\tiny (10)}} & \cellcolor{blue!5} 3.28{\tiny (10)} & \cellcolor{red!5} 63.91{\tiny (10)} & \cellcolor{blue!5} 50.45{\tiny (10)} & \cellcolor{red!5} 47.19{\tiny (10)} & \cellcolor{blue!5} 38.43{\tiny (10)} & \multicolumn{1}{>{\columncolor{red!5}}c|} {32.23{\tiny (10)}} & \cellcolor{blue!5} 46.44{\tiny (10)} \\ \hline

\end{tabular}

}
\vspace*{0.15cm}
\caption{Detailed results on SEEDBench. Five tasks are provided here: Scene Understanding, Instance Identity, Instance Location, Instance Attribute, and Instances Counting. The "Avg." column denotes the average performance across these tasks. Each row contains a relative ranking among all models in the corresponding column (a small number in parentheses), and this ranking is also highlighted with color. }
\label{tab:seedbench}
\end{table*}

\section{OODCV Detailed Result}
\label{app:oodcv}

OODCV-VQA benchmarks includes 7 OOD scenarios:
\begin{itemize}
    \item Weather
    \item Context
    \item Occlusion
    \item IID
    \item Texture
    \item Shape
    \item Pose
\end{itemize}
First, all models surpass the 90\% coverage rate threshold except only CogAgent-VQA on the Occlusion subset and MoE-LLaVA-Phi2-2.7B on the IID subset. 
Except Occlusion scenario all accuracy values fall within a small range. In the Occlusion scenario, four models (LLaVA-v1.6-13B, Monkey-Chat, InternLM-XComposer2-VL, MoE-LLaVA-Phi2-2.7B) demonstrate near-perfect accuracy, while some other models show accuracy below 70\%.

We also notice a similar situation in terms of SS metric. Moreover, the four mentioned models have Set Sizes close to one on Occlusion scenario, indicating very high certainty of their prediction. We do not observe any other outliers across other scenarios.

\begin{table*}[t]
\centering
\resizebox{\textwidth}{!}{%
\begin{tabular}{l|l|ccccccccc|ccccccccc}
\hline
\multirow{2}{*}{\textbf{VLM}} & \multirow{2}{*}{\textbf{LLM}} & \multicolumn{8}{c|}{\textbf{Coverage Rate (\%)}} & 
\multicolumn{8}{c}{\textbf{Acc (\%)} $\uparrow$} \\ 
\cline{3-18} 
&  & \textbf{W} & \textbf{Con} & \textbf{Occ} & \textbf{IID} & \textbf{Txt} & \textbf{Shape} & \multicolumn{1}{c|}{\textbf{Pose}} & \textbf{Avg.} & \textbf{W} & \textbf{Con} & \textbf{Occ} & \textbf{IID} & \textbf{Txt} &  \textbf{Shape} & \multicolumn{1}{c|}{\textbf{Pose}} & \textbf{Avg.} \\ \hline 
LLaVA-v1.6-13B & Vicuna-13b & 97.21 & 95.03 & 99.33 & 92.89 & 92.38 & 95.18 & 96.26 & 95.47 & \cellcolor{red!35} 75.42{\tiny (4)} & \cellcolor{blue!25} 72.77{\tiny (6)} & \cellcolor{red!45} 99.33{\tiny (2)} & \cellcolor{blue!25} 60.34{\tiny (6)} & \multicolumn{1}{>{\columncolor{red!30}}c|} {69.53{\tiny (5)}} & \cellcolor{blue!40} 75.0{\tiny (3)} & \cellcolor{red!20} 70.05{\tiny (7)} & \cellcolor{blue!30} 74.64{\tiny (5)} \\
Monkey-Chat & Qwen-7b & 93.02 & 95.55 & 99.00 & 93.32 & 93.16 & 95.39 & 93.58 & 94.72 & \cellcolor{red!30} 74.3{\tiny (5)} & \cellcolor{blue!35} 73.3{\tiny (4)} & \cellcolor{red!35} 98.67{\tiny (4)} & \cellcolor{blue!20} 58.62{\tiny (7)} & \multicolumn{1}{>{\columncolor{red!40}}c|} {70.7{\tiny (3)}} & \cellcolor{blue!35} 74.56{\tiny (4)} & \cellcolor{red!25} 70.59{\tiny (6)} & \cellcolor{blue!25} 74.39{\tiny (6)} \\
LLaVA-v1.6-7B & Vicuna-7B & 94.97 & 97.64 & 92.00 & 92.24 & 93.16 & 96.05 & 94.39 & 94.35 & \cellcolor{red!45} 77.09{\tiny (2)} & \cellcolor{blue!40} 75.39{\tiny (3)} & \cellcolor{red!25} 88.0{\tiny (6)} & \cellcolor{blue!50} 64.66{\tiny (1)} & \multicolumn{1}{>{\columncolor{red!15}}c|} {67.58{\tiny (8)}} & \cellcolor{blue!45} 77.63{\tiny (2)} & \cellcolor{red!45} 75.94{\tiny (2)} & \cellcolor{blue!40} 75.18{\tiny (3)} \\
InternLM-XComposer2-VL & InternLM-7b & 91.62 & 94.24 & 99.67 & 90.09 & 92.77 & 94.74 & 93.05 & 93.74 & \cellcolor{red!15} 69.83{\tiny (8)} & \cellcolor{blue!15} 70.16{\tiny (8)} & \cellcolor{red!45} 99.33{\tiny (2)} & \cellcolor{blue!15} 55.17{\tiny (8)} & \multicolumn{1}{>{\columncolor{red!20}}c|} {68.75{\tiny (7)}} & \cellcolor{blue!10} 69.3{\tiny (9)} & \cellcolor{red!10} 67.38{\tiny (9)} & \cellcolor{blue!20} 71.42{\tiny (7)} \\
Yi-VL-6B & Yi-6B & 91.06 & 96.07 & 94.67 & 93.97 & 91.80 & 95.39 & 94.12 & 93.87 & \cellcolor{red!45} 77.09{\tiny (2)} & \cellcolor{blue!35} 73.3{\tiny (4)} & \cellcolor{red!30} 91.33{\tiny (5)} & \cellcolor{blue!35} 62.07{\tiny (4)} & \multicolumn{1}{>{\columncolor{red!45}}c|} {73.44{\tiny (2)}} & \cellcolor{blue!15} 71.05{\tiny (8)} & \cellcolor{red!50} 77.01{\tiny (1)} & \cellcolor{blue!35} 75.04{\tiny (4)} \\
CogAgent-VQA & Vicuna-7B & 96.65 & 94.24 & 89.67 & 91.81 & 94.53 & 96.71 & 93.05 & 93.81 & \cellcolor{red!25} 72.63{\tiny (6)} & \cellcolor{blue!20} 70.68{\tiny (7)} & \cellcolor{red!15} 64.0{\tiny (8)} & \cellcolor{blue!15} 55.17{\tiny (8)} & \multicolumn{1}{>{\columncolor{red!25}}c|} {69.14{\tiny (6)}} & \cellcolor{blue!25} 74.12{\tiny (6)} & \cellcolor{red!15} 68.45{\tiny (8)} & \cellcolor{blue!15} 67.74{\tiny (8)} \\
MobileVLMV2-7B & Vicuna-7B & 96.37 & 95.55 & 90.67 & 93.53 & 94.73 & 94.96 & 93.05 & 94.12 & \cellcolor{red!10} 67.6{\tiny (9)} & \cellcolor{blue!10} 68.06{\tiny (9)} & \cellcolor{red!10} 62.0{\tiny (9)} & \cellcolor{blue!30} 60.78{\tiny (5)} & \multicolumn{1}{>{\columncolor{red!10}}c|} {67.19{\tiny (9)}} & \cellcolor{blue!35} 74.56{\tiny (4)} & \cellcolor{red!30} 71.66{\tiny (5)} & \cellcolor{blue!10} 67.41{\tiny (9)} \\
MoE-LLaVA-Phi2-2.7B & Phi2-2.7B & 93.58 & 91.62 & 99.67 & 86.21 & 92.77 & 94.96 & 94.65 & 93.35 & \cellcolor{red!25} 72.63{\tiny (6)} & \cellcolor{blue!50} 76.44{\tiny (1)} & \cellcolor{red!45} 99.33{\tiny (2)} & \cellcolor{blue!45} 63.36{\tiny (2)} & \multicolumn{1}{>{\columncolor{red!35}}c|} {69.92{\tiny (4)}} & \cellcolor{blue!20} 73.25{\tiny (7)} & \cellcolor{red!35} 75.4{\tiny (4)} & \cellcolor{blue!50} 75.76{\tiny (1)} \\
mPLUG-Owl2 & LLaMA2-7B & 95.81 & 93.19 & 91.00 & 93.53 & 91.02 & 97.37 & 92.51 & 93.49 & \cellcolor{red!50} 78.21{\tiny (1)} & \cellcolor{blue!45} 75.92{\tiny (2)} & \cellcolor{red!20} 82.0{\tiny (7)} & \cellcolor{blue!45} 63.36{\tiny (2)} & \multicolumn{1}{>{\columncolor{red!50}}c|} {74.61{\tiny (1)}} & \cellcolor{blue!50} 78.51{\tiny (1)} & \cellcolor{red!45} 75.94{\tiny (2)} & \cellcolor{blue!45} 75.51{\tiny (2)} \\
Qwen-VL-Chat & Qwen-7b & 92.74 & 93.98 & 95.00 & 94.61 & 92.97 & 92.32 & 92.25 & 93.41 & \cellcolor{red!5} 50.84{\tiny (10)} & \cellcolor{blue!5} 55.5{\tiny (10)} & \cellcolor{red!5} 48.67{\tiny (10)} & \cellcolor{blue!5} 51.72{\tiny (10)} & \multicolumn{1}{>{\columncolor{red!5}}c|} {60.94{\tiny (10)}} & \cellcolor{blue!5} 63.6{\tiny (10)} & \cellcolor{red!5} 59.89{\tiny (10)} & \cellcolor{blue!5} 55.88{\tiny (10)} \\ \hline

\multirow{2}{*}{\textbf{VLM}} & \multirow{2}{*}{\textbf{LLM}} & \multicolumn{8}{c|}{\textbf{Coverage Rate (\%)}} & 
\multicolumn{8}{c}{\textbf{UAcc (\%)} $\uparrow$} \\ 
\cline{3-18} 
&  & \textbf{W} & \textbf{Con} & \textbf{Occ} & \textbf{IID} & \textbf{Txt} & \textbf{Shape} & \multicolumn{1}{c|}{\textbf{Pose}} & \textbf{Avg.} & \textbf{W} & \textbf{Con} & \textbf{Occ} & \textbf{IID} & \textbf{Txt} &  \textbf{Shape} & \multicolumn{1}{c|}{\textbf{Pose}} & \textbf{Avg.} \\ \hline  
LLaVA-v1.6-13B & Vicuna-13b & \cellcolor{red!35} 2.17{\tiny (4)} & \cellcolor{blue!40} 2.34{\tiny (3)} & \cellcolor{red!45} 1.31{\tiny (2)} & \cellcolor{blue!30} 2.6{\tiny (5)} & \multicolumn{1}{>{\columncolor{red!30}}c|} {2.29{\tiny (5)}} & \cellcolor{blue!40} 2.27{\tiny (3)} & \cellcolor{red!25} 2.38{\tiny (6)} & \cellcolor{blue!35} 2.19{\tiny (4)} & \cellcolor{red!40} 88.38{\tiny (3)} & \cellcolor{blue!40} 78.89{\tiny (3)} & \cellcolor{red!45} 196.76{\tiny (2)} & \cellcolor{blue!30} 58.48{\tiny (5)} & \multicolumn{1}{>{\columncolor{red!35}}c|} {80.56{\tiny (4)}} & \cellcolor{blue!40} 83.16{\tiny (3)} & \cellcolor{red!25} 73.03{\tiny (6)} & \cellcolor{blue!35} 94.18{\tiny (4)} \\
Monkey-Chat & Qwen-7b & \cellcolor{red!10} 3.11{\tiny (9)} & \cellcolor{blue!10} 3.37{\tiny (9)} & \cellcolor{red!50} 1.25{\tiny (1)} & \cellcolor{blue!10} 3.36{\tiny (9)} & \multicolumn{1}{>{\columncolor{red!10}}c|} {2.99{\tiny (9)}} & \cellcolor{blue!5} 3.22{\tiny (10)} & \cellcolor{red!10} 3.11{\tiny (9)} & \cellcolor{blue!10} 2.92{\tiny (9)} & \cellcolor{red!10} 62.43{\tiny (9)} & \cellcolor{blue!10} 56.07{\tiny (9)} & \cellcolor{red!50} 200.6{\tiny (1)} & \cellcolor{blue!10} 43.33{\tiny (9)} & \multicolumn{1}{>{\columncolor{red!15}}c|} {63.53{\tiny (8)}} & \cellcolor{blue!10} 58.11{\tiny (9)} & \cellcolor{red!10} 59.09{\tiny (9)} & \cellcolor{blue!20} 77.6{\tiny (7)} \\
LLaVA-v1.6-7B & Vicuna-7B & \cellcolor{red!30} 2.42{\tiny (5)} & \cellcolor{blue!30} 2.55{\tiny (5)} & \cellcolor{red!25} 1.4{\tiny (6)} & \cellcolor{blue!40} 2.49{\tiny (3)} & \multicolumn{1}{>{\columncolor{red!25}}c|} {2.43{\tiny (6)}} & \cellcolor{blue!25} 2.53{\tiny (6)} & \cellcolor{red!20} 2.43{\tiny (7)} & \cellcolor{blue!30} 2.32{\tiny (5)} & \cellcolor{red!30} 86.39{\tiny (5)} & \cellcolor{blue!35} 76.53{\tiny (4)} & \cellcolor{red!25} 166.21{\tiny (6)} & \cellcolor{blue!45} 66.74{\tiny (2)} & \multicolumn{1}{>{\columncolor{red!20}}c|} {69.65{\tiny (7)}} & \cellcolor{blue!35} 78.81{\tiny (4)} & \cellcolor{red!35} 86.24{\tiny (4)} & \cellcolor{blue!30} 90.08{\tiny (5)} \\
InternLM-XComposer2-VL & InternLM-7b & \cellcolor{red!30} 2.42{\tiny (5)} & \cellcolor{blue!35} 2.49{\tiny (4)} & \cellcolor{red!30} 1.37{\tiny (5)} & \cellcolor{blue!15} 3.03{\tiny (8)} & \multicolumn{1}{>{\columncolor{red!35}}c|} {2.23{\tiny (4)}} & \cellcolor{blue!30} 2.52{\tiny (5)} & \cellcolor{red!30} 2.35{\tiny (5)} & \cellcolor{blue!25} 2.35{\tiny (6)} & \cellcolor{red!25} 71.06{\tiny (6)} & \cellcolor{blue!25} 71.96{\tiny (6)} & \cellcolor{red!35} 191.58{\tiny (4)} & \cellcolor{blue!15} 44.77{\tiny (8)} & \multicolumn{1}{>{\columncolor{red!30}}c|} {78.14{\tiny (5)}} & \cellcolor{blue!15} 68.49{\tiny (8)} & \cellcolor{red!15} 72.54{\tiny (8)} & \cellcolor{blue!25} 85.5{\tiny (6)} \\
Yi-VL-6B & Yi-6B & \cellcolor{red!45} 2.01{\tiny (2)} & \cellcolor{blue!25} 2.57{\tiny (6)} & \cellcolor{red!35} 1.34{\tiny (4)} & \cellcolor{blue!45} 2.48{\tiny (2)} & \multicolumn{1}{>{\columncolor{red!50}}c|} {1.94{\tiny (1)}} & \cellcolor{blue!35} 2.36{\tiny (4)} & \cellcolor{red!45} 2.04{\tiny (2)} & \cellcolor{blue!40} 2.11{\tiny (3)} & \cellcolor{red!50} 104.24{\tiny (1)} & \cellcolor{blue!30} 74.29{\tiny (5)} & \cellcolor{red!30} 178.18{\tiny (5)} & \cellcolor{blue!40} 62.84{\tiny (3)} & \multicolumn{1}{>{\columncolor{red!50}}c|} {99.58{\tiny (1)}} & \cellcolor{blue!30} 75.34{\tiny (5)} & \cellcolor{red!50} 97.27{\tiny (1)} & \cellcolor{blue!45} 98.82{\tiny (2)} \\
CogAgent-VQA & Vicuna-7B & \cellcolor{red!15} 2.63{\tiny (8)} & \cellcolor{blue!20} 2.58{\tiny (7)} & \cellcolor{red!15} 2.2{\tiny (8)} & \cellcolor{blue!25} 2.76{\tiny (6)} & \multicolumn{1}{>{\columncolor{red!25}}c|} {2.43{\tiny (6)}} & \cellcolor{blue!20} 2.55{\tiny (7)} & \cellcolor{red!35} 2.34{\tiny (4)} & \cellcolor{blue!20} 2.5{\tiny (7)} & \cellcolor{red!20} 70.51{\tiny (7)} & \cellcolor{blue!20} 68.26{\tiny (7)} & \cellcolor{red!15} 72.03{\tiny (8)} & \cellcolor{blue!20} 49.18{\tiny (7)} & \multicolumn{1}{>{\columncolor{red!25}}c|} {71.48{\tiny (6)}} & \cellcolor{blue!25} 74.15{\tiny (6)} & \cellcolor{red!30} 77.19{\tiny (5)} & \cellcolor{blue!15} 68.97{\tiny (8)} \\
MobileVLMV2-7B & Vicuna-7B & \cellcolor{red!20} 2.58{\tiny (7)} & \cellcolor{blue!15} 2.74{\tiny (8)} & \cellcolor{red!10} 2.53{\tiny (9)} & \cellcolor{blue!20} 2.97{\tiny (7)} & \multicolumn{1}{>{\columncolor{red!15}}c|} {2.73{\tiny (8)}} & \cellcolor{blue!15} 2.59{\tiny (8)} & \cellcolor{red!15} 2.56{\tiny (8)} & \cellcolor{blue!15} 2.67{\tiny (8)} & \cellcolor{red!15} 66.57{\tiny (8)} & \cellcolor{blue!15} 63.05{\tiny (8)} & \cellcolor{red!10} 60.17{\tiny (9)} & \cellcolor{blue!25} 50.93{\tiny (6)} & \multicolumn{1}{>{\columncolor{red!10}}c|} {61.1{\tiny (9)}} & \cellcolor{blue!20} 73.57{\tiny (7)} & \cellcolor{red!20} 72.79{\tiny (7)} & \cellcolor{blue!10} 64.03{\tiny (9)} \\
MoE-LLaVA-Phi2-2.7B & Phi2-2.7B & \cellcolor{red!40} 2.05{\tiny (3)} & \cellcolor{blue!50} 1.99{\tiny (1)} & \cellcolor{red!40} 1.33{\tiny (3)} & \cellcolor{blue!50} 2.1{\tiny (1)} & \multicolumn{1}{>{\columncolor{red!45}}c|} {2.09{\tiny (2)}} & \cellcolor{blue!45} 2.17{\tiny (2)} & \cellcolor{red!50} 2.03{\tiny (1)} & \cellcolor{blue!50} 1.97{\tiny (1)} & \cellcolor{red!35} 87.25{\tiny (4)} & \cellcolor{blue!50} 96.69{\tiny (1)} & \cellcolor{red!40} 194.95{\tiny (3)} & \cellcolor{blue!50} 75.32{\tiny (1)} & \multicolumn{1}{>{\columncolor{red!40}}c|} {85.13{\tiny (3)}} & \cellcolor{blue!45} 85.34{\tiny (2)} & \cellcolor{red!45} 95.1{\tiny (2)} & \cellcolor{blue!50} 102.83{\tiny (1)} \\
mPLUG-Owl2 & LLaMA2-7B & \cellcolor{red!50} 1.99{\tiny (1)} & \cellcolor{blue!45} 2.16{\tiny (2)} & \cellcolor{red!20} 1.57{\tiny (7)} & \cellcolor{blue!35} 2.53{\tiny (4)} & \multicolumn{1}{>{\columncolor{red!40}}c|} {2.2{\tiny (3)}} & \cellcolor{blue!50} 2.1{\tiny (1)} & \cellcolor{red!40} 2.11{\tiny (3)} & \cellcolor{blue!45} 2.1{\tiny (2)} & \cellcolor{red!45} 98.95{\tiny (2)} & \cellcolor{blue!45} 89.59{\tiny (2)} & \cellcolor{red!20} 136.49{\tiny (7)} & \cellcolor{blue!35} 62.54{\tiny (4)} & \multicolumn{1}{>{\columncolor{red!45}}c|} {86.95{\tiny (2)}} & \cellcolor{blue!50} 93.42{\tiny (1)} & \cellcolor{red!40} 92.0{\tiny (3)} & \cellcolor{blue!40} 94.28{\tiny (3)} \\
Qwen-VL-Chat & Qwen-7b & \cellcolor{red!5} 3.57{\tiny (10)} & \cellcolor{blue!5} 3.77{\tiny (10)} & \cellcolor{red!5} 2.87{\tiny (10)} & \cellcolor{blue!5} 3.76{\tiny (10)} & \multicolumn{1}{>{\columncolor{red!5}}c|} {3.59{\tiny (10)}} & \cellcolor{blue!10} 3.2{\tiny (9)} & \cellcolor{red!5} 3.33{\tiny (10)} & \cellcolor{blue!5} 3.44{\tiny (10)} & \cellcolor{red!5} 34.96{\tiny (10)} & \cellcolor{blue!5} 36.15{\tiny (10)} & \cellcolor{red!5} 41.85{\tiny (10)} & \cellcolor{blue!5} 33.74{\tiny (10)} & \multicolumn{1}{>{\columncolor{red!5}}c|} {41.75{\tiny (10)}} & \cellcolor{blue!5} 48.85{\tiny (10)} & \cellcolor{red!5} 44.38{\tiny (10)} & \cellcolor{blue!5} 40.24{\tiny (10)} \\ \hline

\end{tabular}

}
\vspace*{0.15cm}
\caption{Detailed results on OODCV. Seven OOD scenarios are provided here. The "Avg." column denotes the average performance across the seven scenarios. Each row contains a relative ranking among all models in the corresponding column (a small number in parentheses), and this ranking is also highlighted with color.}
\label{tab:OODCV}
\end{table*}

\section{Rate of Predicted E or F}
\label{app:ef}

\begin{table*}[h]
\centering
\resizebox{\textwidth}{!}{%
\begin{tabular}{l|cccccc|cccccc}
\hline
\multirow{2}{*}{\textbf{VLM}}  & \multicolumn{6}{c|}{\textbf{E Rate (\%)}} & 
\multicolumn{6}{c}{\textbf{F Rate (\%)} $\uparrow$} \\ 
\cline{2-13} 
& \textbf{MMB} & \textbf{OOD} & \textbf{SQA} & \textbf{SB} & \multicolumn{1}{c|}{\textbf{AI2D}} & \textbf{Avg.} & \textbf{MMB}  & \textbf{OOD} & \textbf{SQA} & \textbf{SB} & \multicolumn{1}{c|}{\textbf{AI2D}} & \textbf{Avg.} \\ \hline 
MoE-LLaVA-Phi2-2.7B & 0.41 & 0.00 & 0.05 & 0.17 & 0.08 & 0.14 & 0.32 & 0.00 & 2.93 & 0.10 & 0.15 & 0.70 \\
MoE-LLaVA-Qwen-1.8B & 0.09 & 0.00 & 0.00 & 0.28 & 0.13 & 0.10 & 0.69 & 3.23 & 2.43 & 0.41 & 0.94 & 1.54 \\
MoE-LLaVA-StableLM-1.6B & 1.37 & 0.00 & 1.06 & 0.17 & 0.26 & 0.57 & 1.51 & \textbf{27.14} & 3.95 & 1.19 & 1.56 & 7.07 \\
MobileVLM-1.7B & 0.00 & 0.00 & 0.00 & 0.00 & 0.00 & 0.00 & 0.00 & 0.00 & 0.00 & 0.00 & 0.00 & 0.00 \\
MobileVLM-3B & 0.00 & 0.00 & 0.00 & 0.00 & 0.00 & 0.00 & 0.09 & 0.00 & 0.10 & 0.04 & 0.09 & 0.06 \\
MobileVLMV2-1.7B & 0.05 & 0.00 & 0.05 & 0.00 & 0.00 & 0.02 & 0.32 & 0.00 & 0.00 & 0.00 & 0.00 & 0.06 \\
MobileVLMV2-3B & 0.14 & 0.00 & 0.91 & 0.04 & 0.52 & 0.32 & 0.05 & 0.00 & 0.10 & 0.30 & 0.63 & 0.21 \\
MobileVLMV2-7B & 1.55 & 0.35 & 1.62 & 0.39 & 1.56 & 1.10 & 1.05 & 2.81 & 2.28 & 1.33 & 2.27 & 1.95 \\
Monkey & 4.89 & \textbf{20.32} & 3.34 & 10.21 & 4.27 & 8.61 & 0.87 & \textbf{43.46} & 5.36 & 12.25 & 6.41 & 13.67 \\
Monkey-Chat & 0.37 & 0.00 & 0.56 & 0.74 & 1.15 & 0.56 & 1.19 & 0.49 & 7.69 & 2.42 & 2.97 & 2.95 \\
Qwen-VL & 6.53 & \textbf{45.22} & 3.79 & 23.04 & 6.49 & 17.02 & 0.09 & 0.00 & 0.66 & 0.13 & 0.40 & 0.26 \\
Qwen-VL-Chat & 3.56 & \textbf{19.55} & 3.59 & 9.30 & 6.42 & 8.49 & 0.00 & 0.00 & 0.05 & 0.01 & 0.18 & 0.05 \\
Yi-VL-34B & 0.23 & 0.00 & 0.00 & 0.01 & 0.00 & 0.05 & 0.37 & 0.00 & 0.96 & 0.18 & 0.12 & 0.33 \\
Yi-VL-6B & 2.19 & 0.00 & 0.71 & 0.34 & 0.72 & 0.79 & 1.46 & 0.00 & 4.40 & 0.34 & 0.65 & 1.37 \\
CogAgent-VQA & 0.69 & 0.84 & 0.91 & 0.39 & 0.77 & 0.72 & 0.00 & 0.00 & 0.05 & 0.00 & 0.12 & 0.03 \\
CogVLM-Chat & 3.38 & 5.91 & 0.61 & 0.80 & 0.90 & 2.32 & 0.00 & 0.00 & 0.00 & 0.00 & 0.01 & 0.00 \\
CogVLM-Grounding-Generalist & 0.00 & 0.00 & 0.15 & 0.00 & 0.08 & 0.05 & 0.00 & 0.00 & 0.00 & 0.03 & 0.23 & 0.05 \\
InternLM-XComposer2 & 11.33 & \textbf{20.53} & 5.67 & 14.12 & 6.01 & 11.53 & 1.83 & 2.88 & 3.19 & 0.34 & 0.67 & 1.78 \\
InternLM-XComposer2-VL & 8.54 & 2.18 & 6.93 & 8.78 & 3.26 & 5.94 & 0.05 & 0.00 & 0.15 & 0.04 & 0.25 & 0.10 \\
LLaVA-v1.5-Vicuna-13B & 1.14 & 0.35 & 1.37 & 0.41 & 1.07 & 0.87 & 0.69 & 0.70 & 1.97 & 0.44 & 0.92 & 0.94 \\
LLaVA-v1.5-Vicuna-7B & 0.37 & 0.07 & 1.42 & 0.06 & 0.06 & 0.39 & 3.61 & 6.33 & 6.02 & 3.67 & 1.81 & 4.29 \\
LLaVA-v1.6-34B & 0.23 & 0.00 & 0.00 & 0.00 & 0.00 & 0.05 & 0.41 & 0.00 & 0.71 & 0.21 & 0.15 & 0.30 \\
LLaVA-v1.6-Mistral-7B & 0.91 & 0.14 & 0.46 & 0.22 & 0.03 & 0.35 & 0.00 & 0.00 & 0.61 & 0.14 & 0.09 & 0.17 \\
LLaVA-v1.6-Vicuna-13B & 1.01 & 0.77 & 1.82 & 0.48 & 0.00 & 0.82 & 0.09 & 0.00 & 0.00 & 0.03 & 0.00 & 0.02 \\
LLaVA-v1.6-Vicuna-7B & 0.64 & 0.14 & 2.33 & 0.17 & 0.01 & 0.66 & 0.09 & 0.35 & 0.15 & 0.08 & 0.03 & 0.14 \\
mPLUG-Owl2 & 0.09 & 0.00 & 0.00 & 0.13 & 0.01 & 0.05 & 1.96 & 0.28 & 1.97 & 1.85 & 0.83 & 1.38 \\ \hline

\end{tabular}

}
\vspace*{0.15cm}
\caption{ The ratio of test instances for which the predicted answer is option E ("I don’t know") or option F ("None
of the above"). Note that neither of them corresponds to the ground truth answer }
\label{tab:ef}
\end{table*}

Here, we analyze the proportion of model predictions that correspond to answers E and F. However, none of these options are correct; there was a significant proportion of cases for some models where these options were selected. These results are provided in Table ~\ref{tab:ef}.

Among the datasets, the one that stands out is OODCV-VQA. Five models (Monkey, Qwen-VL-Chat, Qwen-VL, internlm-xcomposer2-7b and CogAgent-VQA) have E rate higher than 10\%. We checked it and ensured that Qwen models answered with only option E  and finished their answers after that. However, other models demonstrate an E Rate equal to or close to zero on the OODCV dataset. The mentioned models also demonstrate a high E rate on other benchmarks (especially in SEEDBench), sometimes exceeding the value of 10\%, but still lower than on OOD.

Only two models demonstrate a very high F rate: MoE-LLaVA-StableLM-1.6B and Monkey. Thus, Monkey selects one of the additional incorrect answers in almost 64\% of cases when evaluated in OODCV. This demonstrates the importance of testing models on out-of-distribution data since it helps identify their vulnerabilities better. Meanwhile, the Monkey-Chat model is fine with this problem. It can also be noted that the observed phenomenon may be related to the language model Qwen, which is common to both Qwen-VL and Monkey.

\end{document}